\documentclass[conference]{IEEEtran}

\usepackage[utf8]{inputenc}
\usepackage[T1]{fontenc}
\usepackage{graphicx}
\usepackage{amsmath}
\usepackage{amssymb}
\usepackage{booktabs}
\usepackage{tabularx}
\usepackage{array}
\usepackage[table]{xcolor}
\usepackage[hyphens]{url}
\usepackage[
  colorlinks=true,
  hypertexnames=false,
  linkcolor=blue,  
  citecolor=blue,  
  urlcolor=blue    
]{hyperref}

\makeatletter
\providecommand{\captionof}[1]{\def\@captype{#1}\caption}
\makeatother

\title{Coupled Graph--Policy Distillation for Personalized Medication Safety in Older Adults with Multimorbidity}

\author{
\IEEEauthorblockN{
Zihan Wang\textsuperscript{1,*},
Anglin Liu\textsuperscript{1,*},
Rongyi Wang\textsuperscript{2},
Dantong Li\textsuperscript{3},
Yi Lu\textsuperscript{1},
Siqing Yuan\textsuperscript{1},
Hongxia Xu\textsuperscript{4},\\
Zhongtian Long\textsuperscript{5},
Jintai Chen\textsuperscript{1,\ensuremath{\dagger}}}
\IEEEauthorblockA{\footnotesize
\textsuperscript{1}Artificial Intelligence Thrust,
The Hong Kong University of Science and Technology (Guangzhou)\\
\textsuperscript{2}Faculty of Engineering, University of New South Wales\\
\textsuperscript{3}Medical Big Data Center, Guangdong Provincial People's Hospital,
Southern Medical University\\
\textsuperscript{4}Transvascular Implantation Devices Research Institute,
Zhejiang University\\
\textsuperscript{5}School of Computer Science and Technology, Huazhong University of Science and Technology, Wuhan 430074, China.\\
\textsuperscript{*}Equal contribution.
\textsuperscript{\ensuremath{\dagger}}Corresponding author.}
\textsuperscript{\ensuremath{\dagger}}
Corresponding author: jintaiCHEN@hkust-gz.edu.cn
}

\begin{document}

\maketitle

\begin{abstract}
Large language model (LLM) agents can support medication review
between clinical visits, but safe choices for older adults with
multimorbidity depend on conditions, medications, and geriatric risks
that users may omit. We introduce \textbf{ATLAS}, a coupled graph--policy
distillation framework for patient-adaptive medication safety. ATLAS
structures guideline evidence as a medication-safety graph. Targeted
questions update the patient state and distill relevant relations into
a patient-specific medication conflict graph (PMCG). A risk-first
multi-agent policy uses the PMCG to screen contraindications, assess cautions and monitoring needs, identify safer alternatives, and verify the final medication plan. We also introduce \textbf{GeriMedBench}, an
interactive benchmark that tests safety-critical information
acquisition and evidence-based decision revision. Across a European non-interactive multimorbidity benchmark,
an Asian interactive multimorbidity benchmark, and an Asian
non-interactive cross-guideline benchmark, ATLAS achieves the
strongest complete-decision performance among the compared systems.
On the European non-interactive multimorbidity benchmark, it exceeds
the strongest proprietary LLM baseline by 53.73 points in Strict
Success Rate and 14.63 points in overall safety reasoning score (OSRS), with no unsafe recommendations
under the automated evaluator. A blinded clinician evaluation gives
ATLAS higher mean ratings across all five criteria and flags potentially
unsafe recommendations in one ATLAS case and two Gemini cases.
Our code and homepage are available at: \url{https://github.com/HKUSTGZ-ML4Health-Lab/ATLAS}.
\end{abstract}


\begin{figure}[t]
    \centering
    \includegraphics[width=1\linewidth]{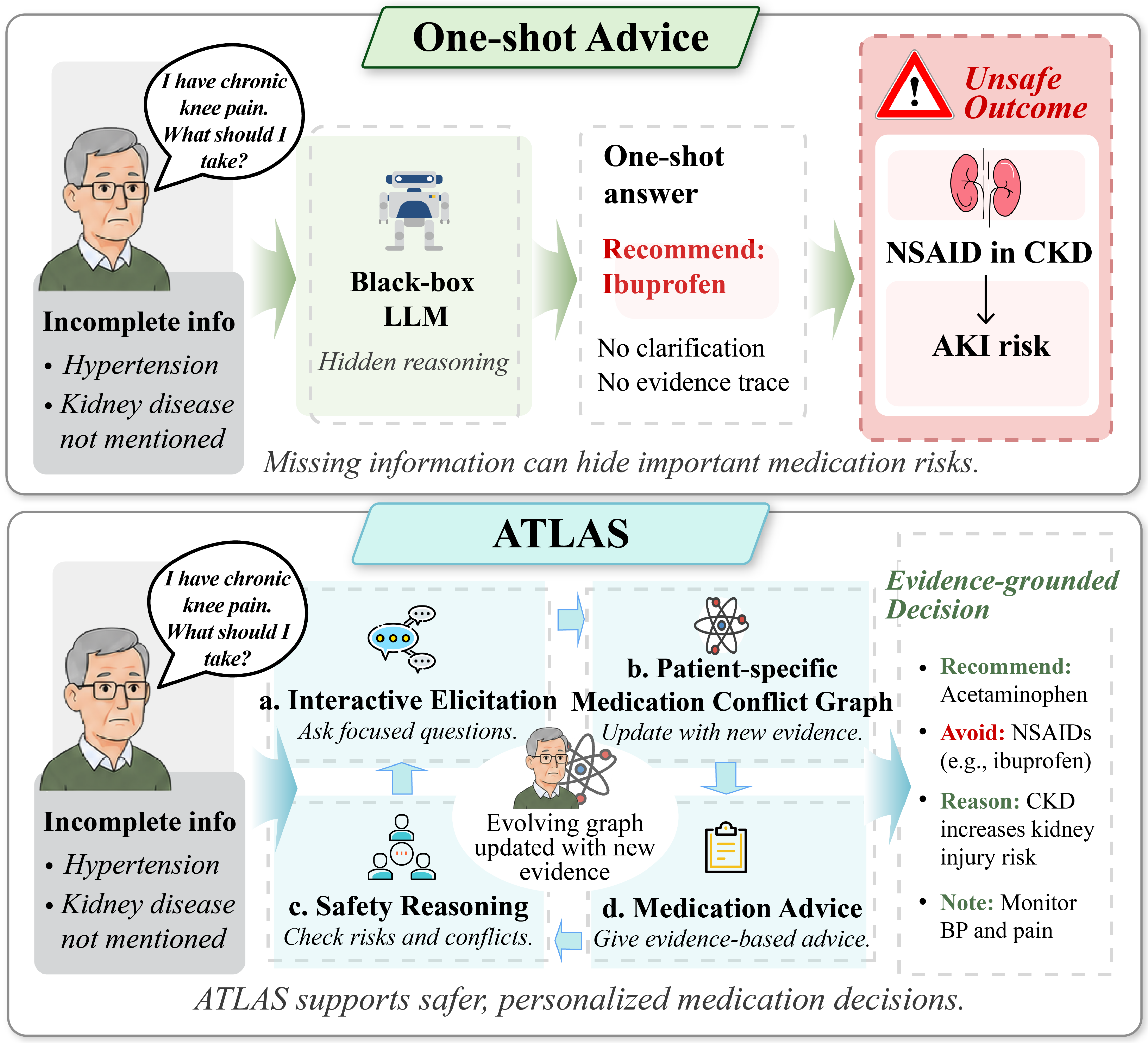}
    \caption{One-shot medication advice versus ATLAS. A black-box LLM may miss unreported risks, while ATLAS elicits missing information, updates the patient-specific medication conflict graph (PMCG), and supports an evidence-grounded decision through safety-first reasoning. }
    \label{fig:motivation}
\end{figure}

\section{Introduction}

\noindent Medication safety for older adults with multimorbidity depends on
clinical facts that an initial query may omit. A user may describe a
symptom or treatment goal without reporting a condition, medication,
or geriatric risk that changes the safety of a candidate treatment.
An agent that treats the first message as a complete patient profile
can produce a plausible but unsafe recommendation. Figure~\ref{fig:motivation}
illustrates this gap. Safe medication decision support must identify
decision-changing information, ask focused questions, and revise the
medication plan as new evidence emerges. ATLAS supports clinician and
pharmacist judgment rather than replacing it \cite{li2024mediq}.

This task poses two linked challenges. Guideline knowledge spans many
medications, conditions, risks, monitoring requirements, and
alternatives, while each patient requires a small and changing subset.
The system must also convert this evidence into an ordered decision
process that prioritizes avoidance, cautions, alternatives, and
verification.

We propose \textbf{ATLAS}, a coupled graph--policy distillation framework.
ATLAS builds a provisional patient-specific medication conflict graph
(PMCG), uses unresolved graph relations to select questions, and
updates the PMCG with each response. A symbolic risk-first policy then
guides conflict screening, caution assessment, alternative selection,
targeted revision, and final verification. The graph identifies which
evidence matters. The policy determines how that evidence changes the
medication plan.

We introduce \textbf{GeriMedBench} to evaluate whether an agent can acquire
missing safety evidence under a query budget and use that evidence to
revise its decision. We evaluate ATLAS on a European non-interactive multimorbidity
benchmark, an Asian interactive multimorbidity benchmark, and an
Asian non-interactive cross-guideline benchmark. On the European non-interactive multimorbidity benchmark, ATLAS
exceeds the strongest proprietary LLM baseline by 53.73 points
in Strict Success Rate and 14.63 points in overall safety reasoning score (OSRS), with no unsafe
recommendations. A blinded clinician evaluation gives ATLAS higher mean ratings across all five criteria. Our contributions are fourfold:
\begin{enumerate}
    \item \textit{Coupled Graph--Policy Distillation.}
    We distill patient-relevant guideline relations into a PMCG and
teacher trajectories into a symbolic risk-first policy executed by
a structured multi-agent system.

    \item \textit{Graph-Guided Interactive Personalization.}
    Unresolved graph dependencies trigger targeted questions, and each response updates the PMCG and selectively revises the safety assessments and medication decisions it supports.

    \item \textit{GeriMedBench.}
    We introduce an interactive benchmark for evaluating safety-critical information acquisition, budgeted questioning, and evidence-grounded, trace-consistent medication decisions.

    \item \textit{Cross-Setting and Cross-Protocol Evaluation.}
We evaluate ATLAS on a European non-interactive multimorbidity
benchmark, an Asian interactive multimorbidity benchmark, and an
Asian non-interactive cross-guideline benchmark.
\end{enumerate}

\section{Related Work}

\noindent\textbf{Drug Recommendation.}
Drug recommendation methods combine molecular structures, DDI knowledge, and patient records. SafeDrug \cite{yang2021safedrug} models DDIs with molecular graphs, while MoleRec \cite{yang2023molerec} captures patient-specific effects of drug substructures. CT-PASMR \cite{ge2025personalized} and EDRMM \cite{liu2025edrmm} improve drug representations through hybrid and multi-granularity modeling, while TraceDR \cite{lin2025tracedr}, DrugRec \cite{sun2022drugrec}, and Hyper-Geri \cite{li2025hypergeri} incorporate knowledge graphs, traceability, external knowledge, or hypergraph learning. These methods assume fixed patient profiles. They do not ask for missing multimorbidity information or revise medication-safety decisions through interaction. 

\noindent\textbf{LLM-Based Clinical Agents.}
LLMs increasingly support clinical decision-making through multi-agent
collaboration. MedAgents \cite{tang2024medagents} uses multidisciplinary
discussion for medical question answering, ClinicalAgent
\cite{yue2024clinicalagent} targets clinical trial tasks, DrAgent
\cite{liu2025dragent} combines clinical tools with recursive learning,
and MedRad \cite{du2025medrad} integrates knowledge engineering with
retrieval-augmented reasoning. Related work also explores MDAgents \cite{kim2024mdagents}. These systems address clinical reasoning and agent collaboration, but they do not center on patient-adaptive medication safety for older adults with multimorbidity. 

\noindent\textbf{Process-Oriented Agent Evaluation.}
Agent evaluation has expanded from final-answer accuracy to interactive reasoning and safety. AgentBench \cite{liu2024agentbench} evaluates agents across interactive environments, and MedicalAgentsBench \cite{shao2025medicalagentsbench} focuses on multi-step clinical reasoning and treatment planning. NOHARM \cite{wu2025noharm}, MATRIX \cite{lim2025matrix}, CSEDB \cite{wang2026csedb}, and CARES \cite{chen2025cares} assess medical harm, hazardous clinical dialogues, safety and efficacy, or adversarial robustness. However, existing benchmarks do not specifically test whether medication agents can identify missing safety-critical information, ask informative questions under a limited budget, and use newly acquired evidence to revise decisions consistently.

\begin{figure*}[!t]
    \centering
    \includegraphics[width=\linewidth]{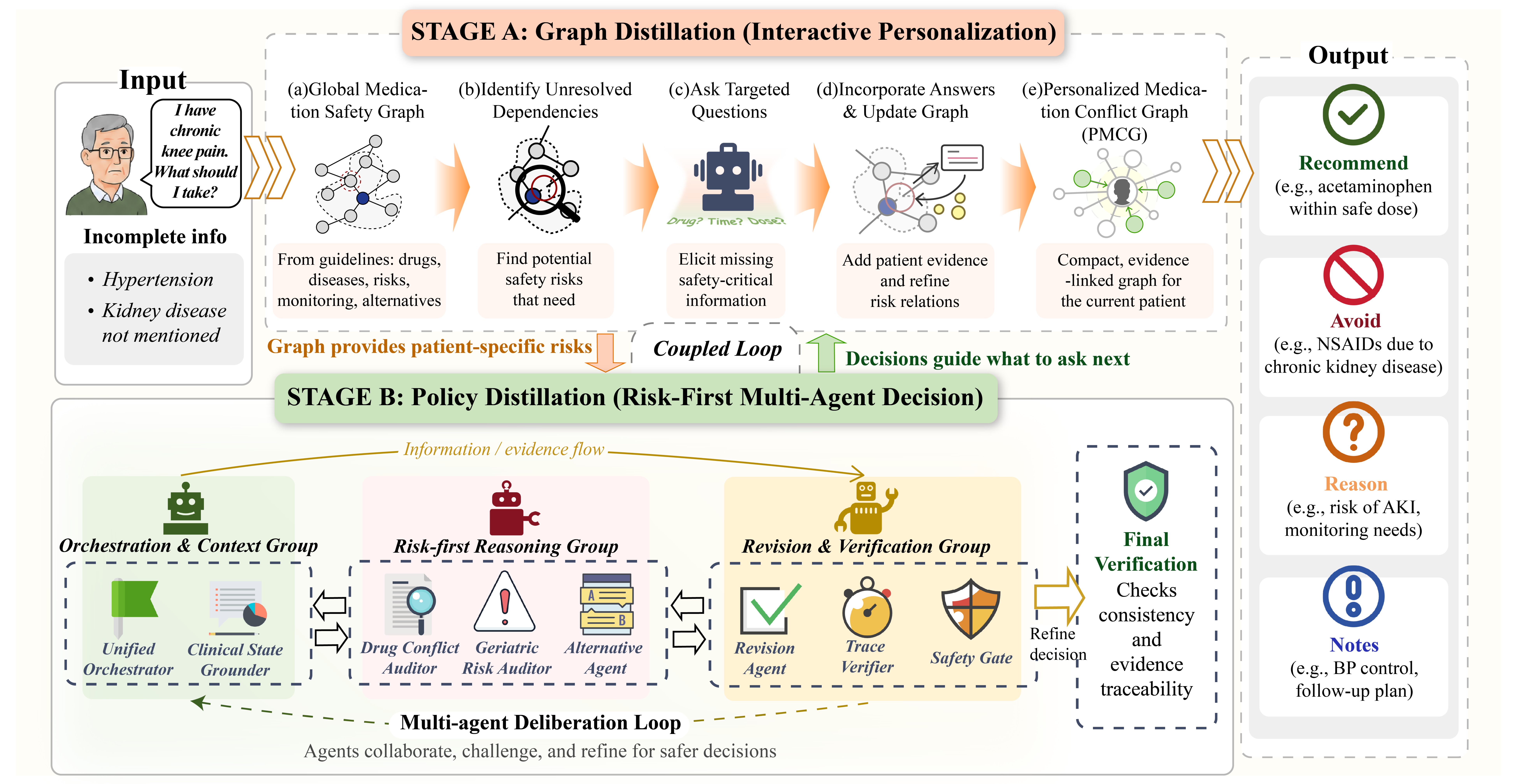}
    \caption{ATLAS architecture. The graph branch updates an evolving
PMCG through targeted questioning. The policy branch executes the
distilled risk-first policy to assess conflicts, risks, alternatives,
and recommendations, then verifies the final medication plan. The
three agent groups correspond to the three functional layers defined
in the text. The output panel illustrates an example report rather than the full
evaluation schema, which contains recommendation, avoidance, caution,
and alternative components. }
    \label{fig:atlas_overview}
\end{figure*}

\section{Method}
\label{sec:method}

ATLAS treats medication decision support as an evidence-based
consultation. It updates a structured patient state, distills relevant
guideline relations into a PMCG, and executes a risk-first policy over
the graph. The PMCG identifies unresolved safety dependencies. The
policy selects questions, revises affected decisions, and verifies the
final medication plan. Figure~\ref{fig:atlas_overview} summarizes graph distillation and
risk-first policy execution.

\subsection{Patient State and Three-Layer Agent Architecture}
\label{sec:patient_state_agent_architecture}

At each dialogue round $t$, ATLAS represents the patient state
with five components:
\begin{equation}
  s_t = (D_t, M_t, A_t, R_t, C_t),
  \label{eq:patient-state}
\end{equation}
Here, $D_t$, $M_t$, $A_t$, $R_t$, and $C_t$ represent the
diagnosed conditions in the multimorbidity profile, current
medications, age and geriatric factors, safety modifiers, and
therapeutic context, respectively. The therapeutic context includes
the presenting problem, treatment goal, and decision setting. ATLAS
treats unreported information as unknown rather than assuming its
absence.

ATLAS groups its agents into three layers. The \emph{Orchestration and Context Layer} contains the Unified Orchestrator and Clinical State
Grounder.  The \emph{Graph Personalization
and Safety Audit Layer} contains
the Drug Conflict Auditor and Geriatric Risk Auditor. The \emph{Decision Synthesis and Verification Layer} contains the Alternative Agent,
Revision Agent, Trace Verifier, and Safety Gate. A shared blackboard
passes patient evidence, PMCG updates, and decision revisions across
the layers.

\subsection{Stage I: Clinical Intake and Treatment Goal Setting}
\label{subsec:clinical-intake}

At the start of each consultation, ATLAS identifies the primary
clinical concern and treatment goal, then records the available
patient-state components. In later rounds, the Clinical State Grounder converts each patient response into a structured state update and retains the original wording for explanation.

ATLAS does not use a fixed questionnaire. It builds a provisional
PMCG and asks only about missing information that could change the
medication decision.

\subsection{Stage II: Progressive PMCG Distillation with Targeted Questioning}
\label{subsec:pmcg-distillation}

The guideline graph links medications, conditions, risk factors,
therapeutic goals, monitoring requirements, alternatives, and
supporting evidence~\cite{zhou2026druganalysis}. ATLAS retains
relations that match the current patient state, removes inapplicable
relations, and marks those that depend on missing information as
unresolved. Active relations support the current decision, while
unresolved relations identify missing information that could still
change the medication plan.

At each round, ATLAS ranks candidate questions first by unresolved
risk severity and then by the number of medication decisions each
question may affect. It uses the question history to break ties.
After the patient responds, ATLAS adds the new evidence to the
patient state and rebuilds the PMCG from the updated state and the
relevant guideline graph:
\begin{equation}
  \begin{array}{l}
    s_{t+1} = s_t \oplus \Phi(q_t, a_t), \\
    G^{\mathrm{p}}_{t+1} = \mathcal{P}(G^{\mathrm{g}}; g, s_{t+1}).
  \end{array}
  \label{eq:pmcg-update}
\end{equation}
Here, \(\Phi\) maps question \(q_t\) and answer \(a_t\) to a structured
state update, while \(\oplus\) integrates the new evidence into the
current patient state. \(G^g\) denotes the guideline graph, \(g\) the
treatment goal, and \(\mathcal{P}\) the personalization operator that
constructs the updated PMCG \(G^p_{t+1}\).

After each PMCG update, the Drug Conflict Auditor reassesses
contraindications and medication--condition conflicts, while the
Geriatric Risk Auditor reassesses age-related risks, caution
requirements, and monitoring needs. These audits determine whether
the new evidence reveals a contraindication, resolves an earlier
concern, introduces a monitoring requirement, or changes the
preferred alternative. ATLAS stops questioning when no unresolved issue can meaningfully change the decision, no informative question remains, or the query
budget runs out. By reconstructing the PMCG from the current
patient state and relevant guideline evidence, ATLAS can apply the same procedure across multimorbidity settings and guideline sources. This coupling works in both directions. Unresolved PMCG relations guide the next
question, while each response updates the graph and limits revisions
to medication decisions affected by the changed relations.

\subsection{Stage III: Consultation-Style Safety-First Decision}
\label{subsec:safety-first-decision}

The current PMCG guides ATLAS to screen contraindications and serious
conflicts before cautions, monitoring needs, and alternatives. ATLAS
gives avoidance priority over treatment preference and removes
high-risk options from the recommendation component. The Alternative
Agent searches guideline evidence when an auditor excludes an option.

ATLAS uses its consultation agents as a multi-agent teacher. The
teacher runs the 39 development cases under
\(K\in\{1,2,3\}\) and produces 117 trajectories. Each trajectory
records questions, patient-state updates, PMCG transitions, risk
assessments, revisions, stopping decisions, and evidence paths.
Policy distillation extracts recurring guideline-consistent
transitions and encodes them in a versioned YAML rule table. The
symbolic policy governs question selection, risk priority,
reconciliation, revision, stopping, and verification. It contains no
case identifiers, evaluation labels, or learned parameters.

During inference, the agents execute this policy over the PMCG. The
Clinical State Grounder applies alias normalization, entity
canonicalization, negation handling, uncertainty handling, and
component-aware parsing. PMCG-linked templates generate questions and
evidence-grounded explanations. The Revision Agent updates decisions
and evidence paths that depend on changed relations. ATLAS invokes no
external LLM during inference. The supplementary material reports the
policy, trajectory schema, teacher configuration, and inference
procedure.

\subsection{Stage IV: Medication Reconciliation and Decision Verification}
\label{subsec:medication_reconciliation}

The Revision Agent resolves overlaps among the recommendation,
avoidance, and caution components using the priority order of
avoidance, caution, and recommendation. It then checks each alternative
against the revised medication components. The Trace Verifier links
each claim to the current PMCG and evidence path. The Safety Gate
checks component consistency and unresolved safety conflicts. A failed
check returns the affected decision to the Revision Agent. ATLAS
releases the medication plan after both checks pass.

\subsection{GeriMedBench: Interactive Medication Safety Benchmark}
\label{sec:gerimedbench}

\begin{figure}[t]
    \centering
    \includegraphics[width=1\linewidth]{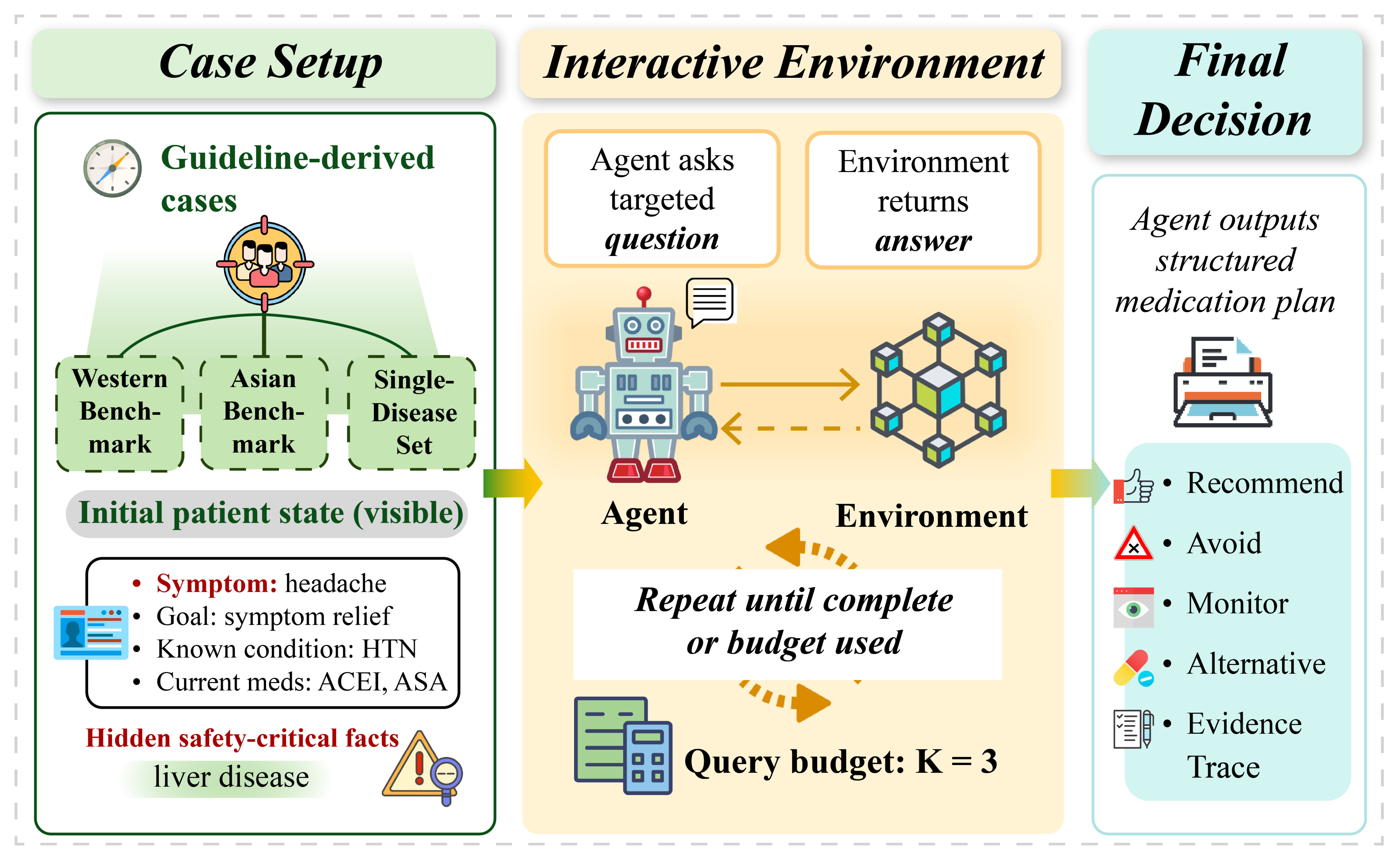}
    \caption{GeriMedBench, an interactive medication-safety benchmark.
The case-setup panel summarizes the evaluation sets, while only the
Asian Multimorbidity Evaluation Set uses the interactive protocol.
Under this protocol, an agent queries an incomplete patient state
with hidden safety facts and revises its final decision under a
budget of \(K=3\). }
    \label{fig:gerimedbench}
\end{figure}

GeriMedBench frames medication safety as an interactive task rather
than a decision based on a complete patient profile. Each case
provides an initial public state, hidden safety-critical facts that
may change the decision, a response environment, and a
guideline-grounded reference. As Figure~\ref{fig:gerimedbench}
shows, the agent interacts with the environment under a limited
query budget before producing a structured medication-safety
decision~\cite{jiang2025medagentbench}.

At round $t$, the system observes the public state $s_t$ and
either asks a question $q_t$ or returns a medication decision $y_t$.
When the system asks a question, the environment returns a patient
answer $a_t$ based on the hidden patient state. The system then updates
the public state using only the newly revealed evidence:
\begin{equation}
  s_{t+1} = s_t \oplus \phi(q_t, a_t).
  \label{eq:benchmark-update}
\end{equation}
Here, $\phi$ extracts the patient information that $a_t$ reveals,
while $\oplus$ adds this information to the public state without
exposing any other hidden facts. The query budget $K$ sets the
maximum number of questions the system can ask.

The interaction ends when the system returns a final decision or
exhausts its budget of $K$ questions. Static benchmarks test whether
a system can reason from a complete patient profile. GeriMedBench
instead tests whether a system can identify decision-relevant missing
information, acquire it efficiently, and use the new evidence to
revise its final decision. The evaluation measures patient-state and decision updates, final medication
safety, and consistency with the evidence trace. Clinical guidelines
ground all patient facts, safety relations, and reference decisions.

\begin{table*}[!t]
\caption{Main results on the Western Multimorbidity Evaluation Set
(\(N=201\)). Values are percentages except OSRS. Best values are bold.
Arrows in the \(\Delta\) row indicate the direction of change relative
to the best non-ATLAS result.}
\centering

\begin{small}
\begin{tabular*}{\textwidth}{
  @{\extracolsep{\fill}}
  lccccccc
  @{}
}
\toprule
\textbf{Method}
& \shortstack{\textbf{Strict}\\\textbf{Success}}
& \shortstack{$\boldsymbol{M_{\mathrm{rec}}}$\\\textbf{F1}}
& \shortstack{$\boldsymbol{M_{\mathrm{avoid}}}$\\\textbf{Recall}}
& \shortstack{$\boldsymbol{M_{\mathrm{caution}}}$\\\textbf{F1}}
& \shortstack{$\boldsymbol{M_{\mathrm{alt}}}$\\\textbf{F1}}
& \shortstack{\textbf{Unsafe}\\\textbf{Rate}}
& \textbf{OSRS} \\
\midrule

\multicolumn{8}{@{}l}{\textbf{\textit{Foundation Models}}} \\
Mistral-Small-3.2-24B
& 2.49 & 78.44 & 87.56 & 42.29 & 9.95 & 12.44 & 63.36 \\
Qwen3-30B-A3B
& 16.42 & 26.37 & 98.01 & 62.19 & 65.17 & 1.99 & 70.85 \\
DeepSeek-R1-Distill-Qwen-32B
& 26.87 & 85.41 & 95.52 & 43.78 & 53.73 & 3.48 & 74.58 \\
MedGemma-27B-Text
& 1.99 & 60.36 & 92.54 & 57.21 & 6.47 & 7.46 & 65.37 \\
GPT-5
& 32.34 & 69.15 & 99.00 & 41.29 & 49.12 & 0.50 & 71.37 \\
Claude Opus 4.6
& 34.83 & 88.56 & 98.01 & 46.77 & 85.32
& \textbf{0.00} & 82.30 \\
Gemini 3.1 Pro Preview
& 38.31 & 63.68 & 99.00 & 70.65 & 80.60
& \textbf{0.00} & 82.58 \\

\multicolumn{8}{@{}l}{\textbf{\textit{Knowledge-Augmented Models}}} \\
Llama-3.3-70B-Instruct
& 0.00 & 79.60 & 96.52 & 51.74 & 1.00
& \textbf{0.00} & 68.82 \\
RotatE
& 0.00 & 30.85 & 13.43 & 20.40 & 0.00 & 62.69 & 28.26 \\

\multicolumn{8}{@{}l}{\textbf{\textit{Multi-Agent Systems}}} \\
MDAgents
& 6.97 & 32.34 & 93.53 & \textbf{79.60} & 7.13
& \textbf{0.00} & 69.42 \\
\midrule

\multicolumn{8}{@{}l}{\textbf{\textit{Ours}}} \\
\rowcolor{gray!15}
\textbf{ATLAS}
& \textbf{92.04}
& \textbf{92.50}
& \textbf{100.00}
& 75.47
& \textbf{92.84}
& \textbf{0.00}
& \textbf{97.21} \\

$\Delta$ vs.\ Best
& {$\uparrow$53.73}
& {$\uparrow$3.94}
& {$\uparrow$1.00}
& {$\downarrow$4.13}
& {$\uparrow$7.52}
& 0.00
& {$\uparrow$14.63} \\
\bottomrule
\end{tabular*}
\end{small}

\label{tab:western_results}
\end{table*}

\begin{table*}[!t]
\caption{GeriMedBench results on the Asian Multimorbidity Evaluation Set
($N=76$, $K=3$). Values are percentages except Agent OSRS. Best values
are bold. The final row reports differences between ATLAS and the best
non-ATLAS result in each column.}
\centering

\begin{small}
\begin{tabular*}{\textwidth}{
  @{\extracolsep{\fill}}
  lccccc
  @{}
}
\toprule
\textbf{Method}
& \shortstack{\textbf{Revision}\\\textbf{Acc.} $\uparrow$}
& \shortstack{\textbf{Final}\\\textbf{Strict} $\uparrow$}
& \shortstack{\textbf{Trace}\\\textbf{Consistency} $\uparrow$}
& \shortstack{\textbf{Unsafe}\\\textbf{Rate} $\downarrow$}
& \shortstack{\textbf{Agent}\\\textbf{OSRS} $\uparrow$} \\
\midrule

\multicolumn{6}{@{}l}{\textbf{\textit{Foundation Models}}} \\
Qwen3-30B-A3B
& 43.69 & 3.95 & 75.00 & 10.53 & 55.03 \\
DeepSeek-R1-Distill-Qwen-32B
& 38.64 & 2.63 & 65.79 & 13.16 & 49.03 \\
MedGemma-27B-Text
& 35.43 & 0.00 & 81.58 & \textbf{0.00} & 53.63 \\
GPT-5
& 40.60 & 2.63 & 81.58 & \textbf{0.00} & 58.98 \\
Claude Opus 4.6
& 40.51 & 5.26 & 81.58 & \textbf{0.00} & 62.87 \\
Gemini 3.1 Pro Preview
& 43.18 & 7.89 & 73.68 & 2.63 & 60.54 \\

\multicolumn{6}{@{}l}{\textbf{\textit{Knowledge-Augmented Models}}} \\
Llama-3.3-70B-Instruct
& 42.11 & 6.58 & 84.21 & 3.95 & 58.63 \\
\midrule

\multicolumn{6}{@{}l}{\textbf{\textit{Ours}}} \\
\rowcolor{gray!15}
\textbf{ATLAS}
& \textbf{44.17}
& \textbf{23.68}
& \textbf{85.53}
& \textbf{0.00}
& \textbf{64.09} \\

$\Delta$ vs.\ Best
& {$\uparrow$0.48}
& {$\uparrow$15.79}
& {$\uparrow$1.32}
& 0.00
& {$\uparrow$1.22} \\
\bottomrule
\end{tabular*}
\end{small}

\label{tab:gerimedbench_results}
\end{table*}

\begin{table*}[!t]
\caption{Blinded clinician evaluation on 40 benchmark-stratified
cases. Scores use a five-point scale and average all reviewer--case
ratings. Unsafe cases use case-level majority judgments.}
\centering

\begin{small}
\begin{tabular*}{\textwidth}{
  @{\extracolsep{\fill}}
  lcccccc
  @{}
}
\toprule
\textbf{Method}
& \textbf{Correct.}
& \textbf{Safety}
& \textbf{Complete.}
& \textbf{Action.}
& \textbf{Evidence}
& \textbf{Unsafe} \\
\midrule

Gemini 3.1 Pro Preview
& 3.77
& 4.12
& 3.67
& 3.85
& 3.80
& 2/40 \\

\rowcolor{gray!15}
\textbf{ATLAS}
& \textbf{4.11}
& \textbf{4.35}
& \textbf{4.01}
& \textbf{4.01}
& \textbf{4.08}
& \textbf{1/40} \\

\bottomrule
\end{tabular*}
\end{small}

\label{tab:expert_evaluation}
\end{table*}

\section{Experiments and Results}
\label{sec:experiments}

We evaluate ATLAS on a European non-interactive multimorbidity
benchmark, an Asian interactive multimorbidity benchmark, and an Asian non-interactive
cross-guideline benchmark. The experiments measure complete medication
decisions, safety, evidence acquisition, targeted revision, component
contributions, and cross-guideline robustness.

\begin{figure*}[!t]
\centering
\begin{minipage}[!t]{0.48\textwidth}
    \centering
    \includegraphics[width=\linewidth]
    {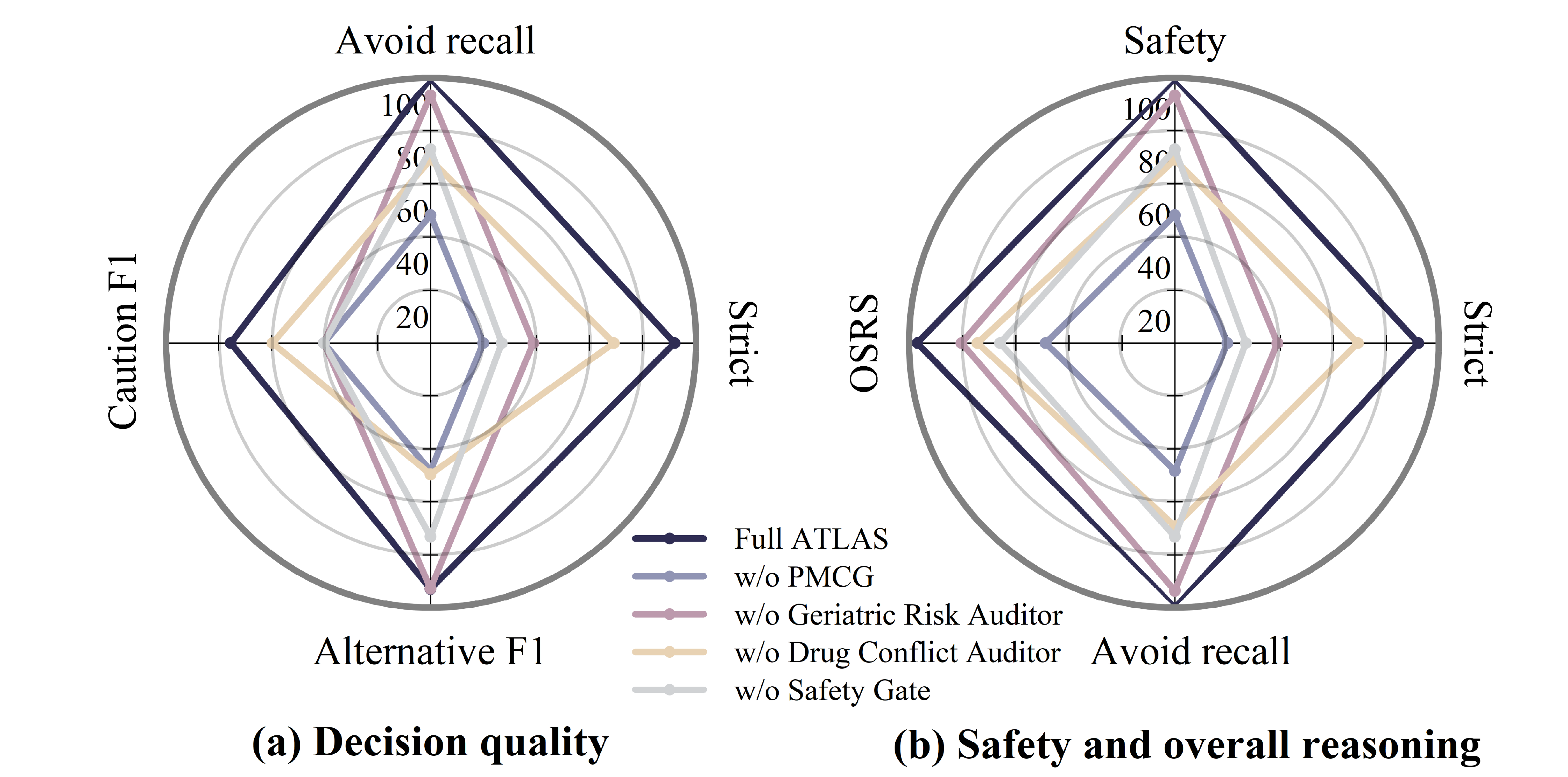}
    \captionof{figure}{Core safety-reasoning ablations. Panel (a) compares
    decision-quality metrics, whereas Panel (b) summarizes safety and overall
    performance. Full ATLAS is compared with four component ablations.
    Safety is defined as $100-\text{Unsafe Recommendation Rate}$.}
    \label{fig:ablation}
\end{minipage}
\hfill
\begin{minipage}[!t]{0.48\textwidth}
    \centering
    \includegraphics[width=\linewidth]
    {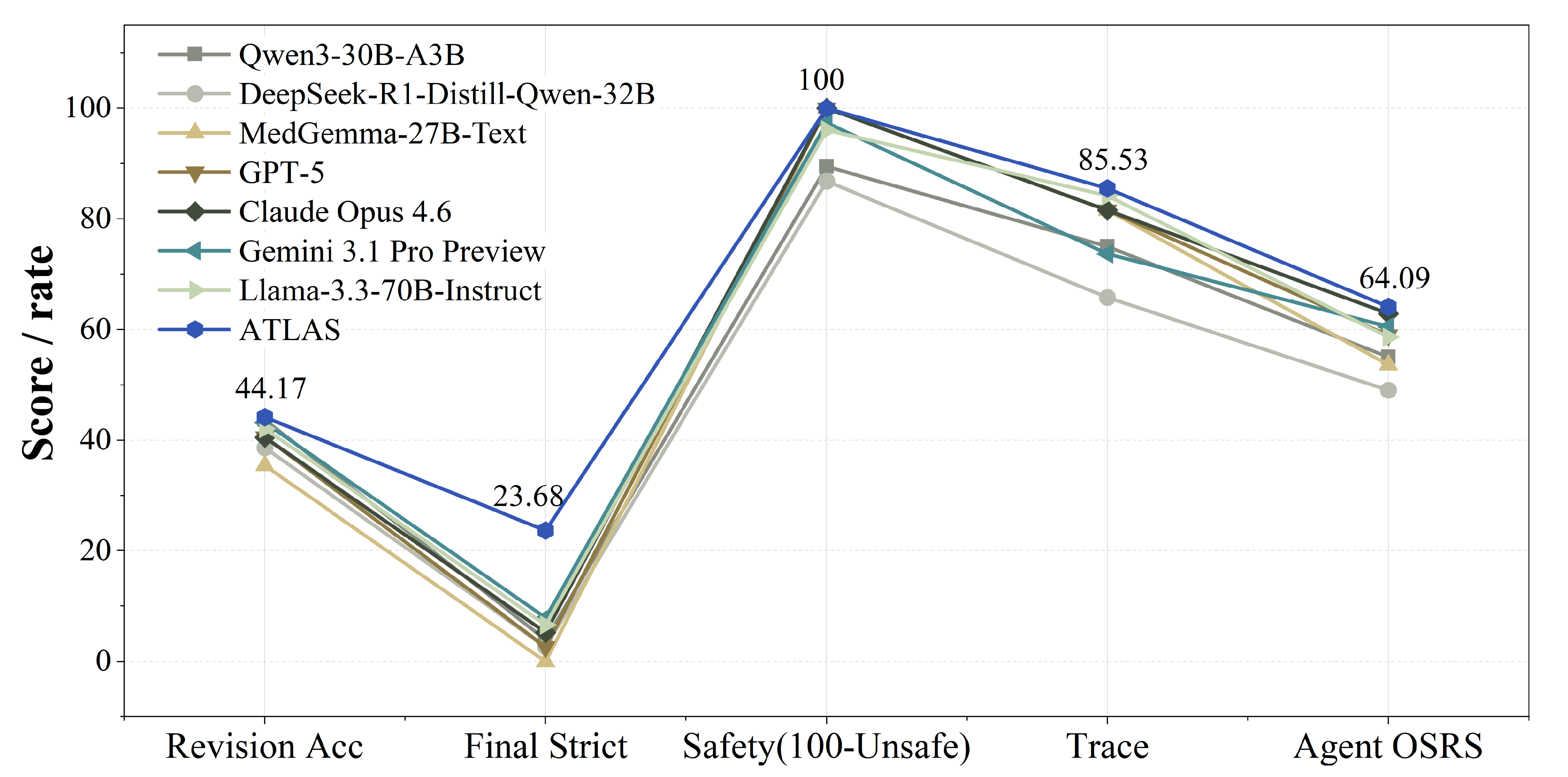}
    \captionof{figure}{GeriMedBench performance on the Asian Multimorbidity
    Evaluation Set ($N=76$, $K=3$). Safety is defined as 100 minus the Unsafe
    Rate. ATLAS is highlighted in blue. Rates are reported as percentages, while Agent OSRS is reported on a 0--100 scale.}
    \label{fig:gerimedbench_results}
\end{minipage}
\end{figure*}

\begin{figure*}[!t]
\centering
\includegraphics[width=\linewidth]{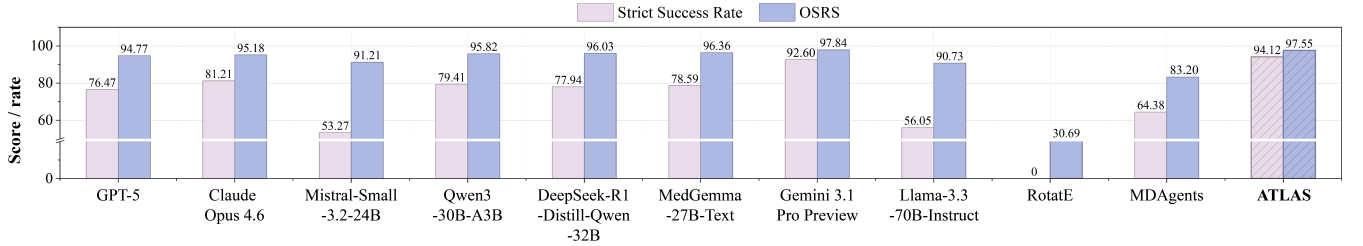}
\caption{Single-disease comparison on the Cross-Regional Single-Disease Generalization Set ($N=612$), 
showing Strict Success Rate and OSRS across methods. Rates are reported as percentages, while OSRS is reported on a 0--100 scale.
}
\label{fig:single_disease}
\end{figure*}

\subsection{Experimental Setup}
\label{sec:experimental_setup}

\paragraph{Datasets and protocols.}
We use a separate 39-case development set for policy construction and
system calibration. No policy or inference component accesses the
evaluation references. Only the frozen offline evaluator reads the test
labels.

The Western Multimorbidity Evaluation Set (\(N=201\)) serves as the
non-interactive multimorbidity benchmark. It is grounded in the AGS
Beers Criteria~\cite{ags2023beers},
STOPP/START version 3~\cite{omahony2023stoppstart},
the FORTA 2024 list~\cite{pazan2025forta},
EURO-FORTA version 2~\cite{pazan2023euroforta}, and the
deprescribing guideline~\cite{quek2025deprescribing}.

The Asian Multimorbidity Evaluation Set (\(N=76\)) serves as the
interactive multimorbidity benchmark under GeriMedBench, where each
agent may ask up to \(K=3\) questions. It is grounded in the Korean
Delphi criteria~\cite{kim2010korean}, the Korean consensus
criteria~\cite{kim2018korean}, and the Japanese guidance on
appropriate medication use in older adults
~\cite{mhlw2018guidance}.

The Cross-Regional Single-Disease Generalization Set (\(N=612\)) serves
as the non-interactive cross-guideline benchmark with
candidate-constrained medication selection. Its reference decisions
are grounded in the ADA guideline for diabetes
~\cite{ada2025pharmacologic}, the AHA/ACC/HFSA guideline for heart
failure~\cite{heidenreich2022hf}, and the KDIGO guideline for chronic
kidney disease~\cite{kdigo2024ckd}.

Clinical guidelines define patient facts, medication relations, and
reference decisions. Each multimorbidity case contains a supported
recommendation and guideline-grounded avoidance, caution, monitoring,
and alternative requirements. These relations define
\(M_{\mathrm{rec}}\), \(M_{\mathrm{avoid}}\),
\(M_{\mathrm{caution}}\), and \(M_{\mathrm{alt}}\). LLMs diversify
scenario wording but do not create safety labels.

\paragraph{Compared methods and evaluation.}
We compare ATLAS with Mistral-Small-3.2-24B,
Qwen3-30B-A3B, DeepSeek-R1-Distill-Qwen-32B,
MedGemma-27B-Text, GPT-5, Claude Opus 4.6, and Gemini 3.1
Pro Preview. Knowledge-augmented baselines include
Llama-3.3-70B-Instruct with BM25 retrieval and RotatE.
MDAgents provides the multi-agent baseline.

All LLM baselines receive the same visible inputs, candidate set,
structured-output instruction, parser, and failure rules. We do not
tune prompts on evaluation cases. No inference method accesses test
references. The supplementary material reports prompts, model
snapshots, decoding settings, retrieval settings, retries, and
invalid-output handling. These comparisons measure end-to-end systems
and do not isolate guideline access from graph personalization and
multi-agent control.

\paragraph{Blinded clinician evaluation.}
Three clinicians independently reviewed 40 benchmark-stratified
cases, with 20 cases from the Western evaluation and 20 from
GeriMedBench. Each reviewer compared one ATLAS output with one
Gemini 3.1 Pro Preview output. We randomized case order and A/B
assignment for each reviewer. We removed model identifiers and
automated scores and presented both outputs through the same
structured template. Reviewers score clinical correctness, medication safety, decision
completeness, actionability, and evidence consistency on a five-point scale \cite{tam2024human}.
They also mark potentially unsafe recommendations and select the preferred
output. We report mean scores, case-level majority judgments, and
Krippendorff's alpha.

\paragraph{Metrics.}
The four medication components are recommendation, avoidance, caution,
and alternative selection. For static evaluation, we report Strict Success,
\(M_{\mathrm{rec}}\) F1, \(M_{\mathrm{avoid}}\) Recall,
\(M_{\mathrm{caution}}\) F1, \(M_{\mathrm{alt}}\) F1,
Unsafe Recommendation Rate, Trace Pass Rate, and OSRS. We use recall for
\(M_{\mathrm{avoid}}\) because a false negative omits a medication
that the patient should avoid. F1 penalizes omissions and unsupported
additions in the other components. For GeriMedBench, we report Revision
Accuracy, Final Strict, Unsafe Recommendation Rate, Trace Consistency,
and Agent OSRS. The supplementary material reports Information Gain
and Query Efficiency.
\begin{equation}
\begin{aligned}
\begin{gathered}
\text{Strict}\\
\text{Success}\\
\text{Rate}
\end{gathered}
&=
\frac{1}{N}\sum_{i=1}^{N}
\mathbb{I}\!\left[
\begin{gathered}
C_{\mathrm{rec},i}
= C_{\mathrm{avoid},i}
= 1,\\
C_{\mathrm{caution},i}
= C_{\mathrm{alt},i}
= 1,\\
C_{\mathrm{unsafe},i}
= C_{\mathrm{trace},i}
= 1
\end{gathered}
\right].
\end{aligned}
\label{eq:strict_success}
\end{equation}

For case \(i\), \(C_{s,i}=1\) when component \(s\) is correct.
\(C_{\mathrm{unsafe},i}=1\) means that the output contains no unsafe
recommendation. Strict Success requires correctness across all six
components. Component metrics separate individual decision errors from
failure of the complete output.

\begin{equation}
\begin{split}
\mathrm{OSRS}={}&0.25R_{\mathrm{avoid}}+0.20R_{\mathrm{rec}}
+0.15R_{\mathrm{caution}} \\
&+0.15R_{\mathrm{alt}}+0.15T+0.10(1-U).
\end{split}
\label{eq:osrs}
\end{equation}

OSRS uses component recall, Trace Pass Rate \(T\), and Unsafe
Recommendation Rate \(U\). We give avoidance recall the largest weight
because missed avoidance relations carry direct safety risk. We
normalize each rate to \([0,1]\) and report OSRS on a 0--100 scale.
We fixed the weights before final evaluation.

\subsection{Results on the Western Multimorbidity Evaluation Set}
\label{subsec:western-results}

Table~\ref{tab:western_results} reports results on the European
non-interactive multimorbidity benchmark. ATLAS achieves 92.04\%
Strict Success and an OSRS of 97.21. It exceeds the best non-ATLAS results by 53.73 points in Strict
Success and 14.63 points in OSRS, with no unsafe recommendations
under the automated evaluator. Strict Success measures joint correctness rather than
an equal gain in every decision component. Several baselines remain
competitive on individual components, but ATLAS completes the full
structured decision more often.

Figure~\ref{fig:ablation} shows that each safety component protects a
different part of the decision. Removing PMCG personalization reduces
Strict Success from 92.04\% to 19.90\%, lowers OSRS from 97.21 to
48.78, and raises Unsafe Rate to 51.74\%. Removing the Geriatric Risk
Auditor lowers \(M_{\mathrm{caution}}\) F1 to 40.30\%. Removing the
Drug Conflict Auditor lowers \(M_{\mathrm{avoid}}\) Recall to 69.15\%
and raises Unsafe Rate to 30.85\%. Removing the Safety Gate reduces Strict Success from 92.04\% to 26.87\% and increases the Unsafe Recommendation Rate from 0\% to 26.87\%.


\subsection{GeriMedBench on the Asian Multimorbidity Evaluation Set}
\label{subsec:asian-results}

Table~\ref{tab:gerimedbench_results} reports the detailed quantitative
results, while Figure~\ref{fig:gerimedbench_results} summarizes the
corresponding performance profiles on the Asian interactive
multimorbidity benchmark. ATLAS achieves the highest Revision Accuracy
of 44.17\%, Final Strict of 23.68\%, Trace Consistency of 85.53\%,
and Agent OSRS of 64.09, with no unsafe recommendations under the
automated evaluator. The Final Strict and Revision Accuracy scores
expose the remaining challenges in information acquisition and
decision revision. Although ATLAS outperforms the compared systems,
these results leave substantial room for improvement.


\subsection{Blinded clinician evaluation.}

Table~\ref{tab:expert_evaluation} complements the automated results
with blinded clinician judgments. ATLAS receives higher mean scores than Gemini 3.1 Pro Preview for clinical correctness, medication safety, decision completeness, actionability, and evidence consistency.
Case-level majority judgments prefer ATLAS in 28 cases, Gemini in
5 cases, and report 7 ties. Reviewers flag potentially unsafe
recommendations in 1 ATLAS case and 2 Gemini cases. Ordinal
Krippendorff's $\alpha$ ranges from 0.33 to 0.71 across the five
criteria. These results support the expert-rated clinical
quality of ATLAS outputs within the reviewed sample. They do not
establish real-world clinical effectiveness.

\subsection{Cross-Regional Single-Disease Generalization Set}
\label{subsec:generalization}

The Asian non-interactive cross-guideline benchmark tests
medication-safety performance outside the multimorbidity settings. As shown in Figure~\ref{fig:single_disease}, ATLAS achieves the highest Strict Success Rate of 94.12\%, exceeding Gemini 3.1 Pro Preview by 1.52 percentage points. ATLAS also records an overall safety reasoning score of 97.55, within 0.29 points of Gemini's best score of 97.84 and above all other methods. These results show that ATLAS delivers the strongest complete-decision performance while maintaining strong overall safety reasoning across guideline sources.


\subsection{Error Analysis and Limitations}
\label{subsec:limitations}

We group failures into missed information, patient-state and PMCG
errors, risk-classification errors, unsuitable alternatives, revision
errors, evidence-alignment errors, and schema errors. On GeriMedBench, we inspect question selection, state updates, PMCG transitions, revisions, and trace grounding. The low Final Strict score exposes errors across the interactive pipeline.

Strict Success and OSRS capture different properties. Strict Success
requires joint correctness. OSRS aggregates component recall, trace
quality, and unsafe outcomes. On the Asian non-interactive cross-guideline benchmark, ATLAS leads Strict Success, while Gemini 3.1 Pro Preview leads OSRS by 0.29 points. No single aggregate metric establishes clinical superiority.

Our comparisons evaluate complete systems and do not separate
structured guideline access from graph personalization and multi-agent
control. The benchmarks use guideline-derived cases, structured
schemas, and candidate-constrained single-disease decisions. They do
not test open-ended prescribing, patient communication, or clinical
outcomes. The blinded expert study covers a small sample and does not
establish clinical effectiveness. More broadly, benchmark performance
alone is insufficient to establish readiness for clinical deployment
~\cite{hager2024limitations}. Future work will extend expert review
and prospective evaluation.

\section{Conclusion}
\label{sec:conclusion}

We present ATLAS, a coupled graph--policy distillation framework for
medication safety in older adults with multimorbidity. ATLAS identifies
missing safety information, updates a PMCG, and revises decisions that
depend on new evidence. A symbolic risk-first policy guides conflict
screening, caution assessment, alternative selection, and final
verification.

We also introduce GeriMedBench to test safety-critical information
acquisition and evidence-based revision. Across a European non-interactive multimorbidity benchmark, an Asian interactive multimorbidity benchmark, and an Asian
non-interactive cross-guideline benchmark, ATLAS achieves the strongest complete-decision performance among
the compared systems and records no unsafe recommendations under
the automated evaluator. A small blinded clinician evaluation gave ATLAS higher mean ratings across all five criteria and fewer majority-flagged unsafe cases than Gemini. These results support graph-guided questioning and evidence-based revision across the tested settings. They do not establish real-world clinical effectiveness.

\bibliographystyle{IEEEtran}
\bibliography{references}


\clearpage
\setcounter{secnumdepth}{0}
\appendix
\section{\textbf{Appendix}}

\section{Contents}

\begin{itemize}
\item \hyperref[app:section-a]{A. Claim--Evidence Map}
\item \hyperref[app:section-b]{B. Benchmark Composition and Data Protocols}
\item \hyperref[app:section-c]{C. ATLAS Implementation Details}
\item \hyperref[app:section-d]{D. GeriMedBench Environment and Evaluation}
\item \hyperref[app:section-e]{E. Baseline Configuration and Output Handling}
\item \hyperref[app:section-f]{F. Additional Quantitative Results}
\item \hyperref[app:section-g]{G. Qualitative Case Study and Error Taxonomy}
\item \hyperref[app:section-h]{H. Blinded Clinician Evaluation}
\item \hyperref[app:section-i]{I. Shared Prompt Templates and Output Schema}
\end{itemize}
\section{A. Claim--Evidence Map}
\label{app:section-a}
\renewcommand{\thetable}{A\arabic{table}}
\setcounter{table}{0}

Table~\ref{tab:supp_a1} gives a direct route from each main-paper claim to the corresponding protocol, artifact, or analysis. It also marks the boundary of each claim, so evidence from one setting is not treated as evidence for another.

\begin{table*}[!t]
\caption{Claim--evidence map for the supplementary material.}
\centering
\begin{small}
\begin{tabular}{@{}p{0.31\textwidth}p{0.43\textwidth}p{0.19\textwidth}@{}}
\toprule
\textbf{Main claim} & \textbf{Primary evidence} & \textbf{Supplementary location} \\
\midrule
ATLAS personalizes medication safety reasoning to incomplete patient states. & Patient-state schema, PMCG relations, agent responsibilities, and the illustrative revision trace. & Sections C and G \\
GeriMedBench evaluates information acquisition and decision revision under a limited query budget. & Environment state, interaction loop, stopping rule, and interactive metrics. & Section D \\
ATLAS improves complete-decision performance across the evaluated settings. & Expanded Western results, GeriMedBench acquisition metrics, and the tabular single-disease comparison. & Section F \\
Different safety modules protect different decision components. & Metric-specific ablation effects and component interpretation. & Section F.4 \\
Clinicians rate ATLAS above Gemini in the reviewed sample. & Review protocol, overall scores, benchmark-stratified scores, and preference counts. & Section H \\
No inference method reads evaluation references. & Reference-access boundaries and shared evaluation controls. & Sections B.3 and E.2 \\
Teacher trajectories are distilled into the symbolic policy used by ATLAS inference. & Trajectory schema, extraction procedure, YAML policy, compilation manifest, and behavioral-equivalence verification. & Sections C.4--C.6 \\
\bottomrule
\end{tabular}
\end{small}

\label{tab:supp_a1}
\end{table*}

\section{B. Benchmark Composition and Data Protocols}
\label{app:section-b}
This section separates dataset construction, information visibility, and output supervision. Table~\ref{tab:supp_b1} identifies the role of each evaluation set, while Tables~\ref{tab:supp_b2} and~\ref{tab:supp_b3} specify what the evaluated systems can observe and what they must return.
\renewcommand{\thetable}{B\arabic{table}}
\setcounter{table}{0}

\subsection{B.1 Evaluation Sets}

The four sets are not interchangeable: each isolates a different stage of development or evaluation. Table~\ref{tab:supp_b1} therefore reports both protocol and primary purpose rather than presenting the sets as a single pooled benchmark.

The evaluation separates development, non-interactive multimorbidity reasoning, interactive multimorbidity reasoning, and cross-guideline single-disease generalization. Table~\ref{tab:supp_b1} summarizes the distinct role of each split.

\begin{table*}[!t]
\caption{Benchmark inventory and evaluation role.}
\centering
\begin{small}
\begin{tabular}{@{}p{0.25\textwidth}cp{0.19\textwidth}p{0.41\textwidth}@{}}
\toprule
\textbf{Set} & \textbf{N} & \textbf{Protocol} & \textbf{Primary purpose} \\
\midrule
Development Set & 39 & Development only & Policy construction and system calibration. No reported test score is computed on this split. \\
Western Multimorbidity Evaluation Set & 201 & Non-interactive & Complete medication-safety decisions from a visible multimorbidity profile. \\
Asian Multimorbidity Evaluation Set & 76 & Interactive, $K=3$ & Question selection, state update, targeted revision, final safety, and trace consistency under GeriMedBench. \\
Cross-Regional Single-Disease Generalization Set & 612 & Non-interactive, candidate-constrained & Cross-guideline generalization outside the multimorbidity settings. \\
\bottomrule
\end{tabular}
\end{small}

\label{tab:supp_b1}
\end{table*}

\subsection{B.2 Case Construction and Label Provenance}

The construction procedure preserves a strict separation between language variation and clinical supervision. This distinction is important when interpreting the benchmark inventory and information boundaries in Tables~\ref{tab:supp_b1} and~\ref{tab:supp_b2}.

The evaluation sets contain guideline-derived, constructed clinical cases rather than identifiable patient records. Guideline sources define patient facts, medication relations, and reference decisions. Each multimorbidity case contains a supported recommendation and guideline-grounded avoidance, caution, monitoring, and alternative requirements. Language models vary surface wording only; they do not create safety labels.

This design supports controlled comparison of medication-safety reasoning over structured guideline evidence. It does not estimate population prevalence, treatment effect, or real-world clinical outcome.

\subsection{B.3 Visible State, Hidden State, and Output Schema}

Table~\ref{tab:supp_b2} contrasts the static and interactive information boundaries at inference time. Table~\ref{tab:supp_b3} then decomposes the required output so that an apparently plausible recommendation cannot compensate for a missed avoidance, caution, alternative, unsafe indicator, or evidence trace.

\begin{table*}[!t]
\caption{Information boundary across static and interactive evaluation.}
\centering
\begin{small}
\begin{tabular}{@{}p{0.18\textwidth}p{0.35\textwidth}p{0.35\textwidth}@{}}
\toprule
\textbf{Element} & \textbf{Non-interactive sets} & \textbf{GeriMedBench} \\
\midrule
Visible patient state & Complete benchmark input provided at inference. & Initial public state with selected safety-critical facts omitted. \\
Hidden safety facts & Not applicable. & Held by the response environment and revealed only through a relevant question. \\
Query action & Not available. & The agent may ask up to $K=3$ targeted questions. \\
Candidate medications & Provided according to the benchmark protocol. & Provided in the public case and held fixed across methods. \\
Final output & Recommendation, avoidance, caution, alternative, unsafe indicator, and evidence trace. & The same structured medication decision after interaction. \\
Reference access & Frozen offline evaluator only. & Environment uses hidden state for answers; final references remain evaluator-only. \\
\bottomrule
\end{tabular}
\end{small}

\label{tab:supp_b2}
\end{table*}

\begin{table*}[!t]
\caption{Structured output components and their principal failure modes.}
\centering
\begin{small}
\begin{tabular}{@{}p{0.17\textwidth}p{0.45\textwidth}p{0.27\textwidth}@{}}
\toprule
\textbf{Component} & \textbf{Required content} & \textbf{Primary failure mode} \\
\midrule
Recommendation ($M_{\mathrm{rec}}$) & Supported medication option and clinically appropriate use context. & Omission or unsupported addition. \\
Avoidance ($M_{\mathrm{avoid}}$) & Medication that should not be used under the case-specific safety evidence. & Missed unsafe option. \\
Caution ($M_{\mathrm{caution}}$) & Monitoring, dose, duration, or conditional-use requirement. & Missing or unsupported caution. \\
Alternative ($M_{\mathrm{alt}}$) & Safer or guideline-supported substitute when a candidate is excluded. & Missing or unsuitable alternative. \\
Unsafe indicator ($U$) & Whether the final plan contains an unsafe recommendation. & Unsafe recommendation remains in the final output. \\
Evidence trace & Link from each claim to patient evidence and guideline-grounded reasoning. & Unsupported or internally inconsistent trace. \\
\bottomrule
\end{tabular}
\end{small}

\label{tab:supp_b3}
\end{table*}

\section{C. ATLAS Implementation Details}
\label{app:section-c}
This section makes the implementation boundary explicit from patient-state representation through frozen-policy execution. Tables~\ref{tab:supp_c1}--\ref{tab:supp_c4} document the graph relations, agent responsibilities, teacher trajectories, and freeze checks that support the reported ATLAS configuration.
\renewcommand{\thetable}{C\arabic{table}}
\setcounter{table}{0}

\subsection{C.1 Patient State and PMCG Representation}

Table~\ref{tab:supp_c1} summarizes the PMCG relation types and shows how each relation affects a medication decision. The separation between support, conflict, caution, alternative, evidence, and unresolved-dependency relations allows updates to remain localized to the affected claims.

At dialogue round $t$, ATLAS represents the patient state as $s_t=(D_t,M_t,A_t,R_t,C_t)$, where the components encode diagnosed conditions, current medications, age and geriatric factors, safety modifiers, and therapeutic context. The system retains unknown values as unknown rather than converting missing information into negative evidence.

\begin{table*}[!t]
\caption{Core PMCG relation types used for personalization and verification.}
\centering
\begin{small}
\begin{tabular}{@{}p{0.17\textwidth}p{0.19\textwidth}p{0.16\textwidth}p{0.36\textwidth}@{}}
\toprule
\textbf{PMCG relation} & \textbf{Source node} & \textbf{Target node} & \textbf{Decision role} \\
\midrule
Recommendation support & Condition, goal, or therapeutic context & Medication candidate & Supports inclusion in $M_{\mathrm{rec}}$ when no higher-priority risk overrides it. \\
Avoidance conflict & Condition, medication, or safety modifier & Medication candidate & Excludes the candidate and activates $M_{\mathrm{avoid}}$. \\
Caution or monitoring dependency & Age factor, organ function, treatment duration, or comorbidity & Medication candidate & Adds a condition, monitoring need, or dose/duration restriction. \\
Alternative relation & Excluded or unsuitable candidate & Safer substitute & Populates $M_{\mathrm{alt}}$ after risk screening. \\
Evidence relation & Guideline source or patient fact & Decision claim & Supports trace verification and final explanation. \\
Unresolved dependency & Missing patient fact & Potential decision change & Triggers a targeted question when the answer can change the plan. \\
\bottomrule
\end{tabular}
\end{small}

\label{tab:supp_c1}
\end{table*}

\subsection{C.2 Agent Responsibilities}

Table~\ref{tab:supp_c2} maps each functional layer to its agents and responsibilities. The decomposition also clarifies which components can revise the patient state, personalize the graph, audit risk, or reconcile a final decision.

\begin{table*}[!t]
\caption{Functional decomposition of the ATLAS agents.}
\centering
\begin{small}
\begin{tabular}{@{}p{0.20\textwidth}p{0.20\textwidth}p{0.49\textwidth}@{}}
\toprule
\textbf{Layer} & \textbf{Agent} & \textbf{Responsibility} \\
\midrule
Orchestration and Context & Unified Orchestrator & Coordinates the consultation, maintains the blackboard, orders agent calls, and enforces the stopping rule. \\
Orchestration and Context & Clinical State Grounder & Normalizes aliases, canonicalizes entities, handles negation and uncertainty, and converts responses into structured state updates. \\
Graph Personalization and Safety Audit & Drug Conflict Auditor & Checks medication--condition conflicts and contraindications and updates avoidance decisions. \\
Graph Personalization and Safety Audit & Geriatric Risk Auditor & Checks age-related risk, organ-function dependencies, cautions, and monitoring needs. \\
Decision Synthesis and Verification & Alternative Agent & Searches guideline evidence for a safer substitute when an option is excluded. \\
Decision Synthesis and Verification & Revision Agent & Updates only the decision components affected by changed PMCG relations and resolves component overlaps. \\
Decision Synthesis and Verification & Trace Verifier & Links each final claim to the current PMCG and evidence path. \\
Decision Synthesis and Verification & Safety Gate & Rejects internally inconsistent plans or plans with unresolved safety conflicts. \\
\bottomrule
\end{tabular}
\end{small}

\label{tab:supp_c2}
\end{table*}

\subsection{C.3 Question Ranking, Revision, and Stopping}

The ordered procedure below links question selection to unresolved PMCG relations rather than to a fixed questionnaire. Its stopping rule prevents additional questions once the remaining uncertainty cannot change a medication component.

ATLAS ranks candidate questions by unresolved risk severity and then by the number of medication decisions that the answer may affect. It uses the question history to break ties. Each answer updates the structured patient state, rebuilds the relevant PMCG subgraph, and triggers a targeted re-audit of the affected decisions.

\begin{enumerate}
\item Build a provisional PMCG from the current patient state and relevant guideline graph.
\item Identify unresolved relations whose resolution may change recommendation, avoidance, caution, or alternative decisions.
\item Rank questions by risk severity, decision coverage, and question history.
\item Integrate the revealed evidence and rebuild the personalized graph.
\item Re-run the affected safety audits and revise only dependent decision components.
\item Stop when no unresolved issue can change the decision, no informative question remains, or the query budget is exhausted.
\end{enumerate}

\subsection{C.4 Teacher Trajectories and Policy Distillation}

Table~\ref{tab:supp_c3} lists every field retained in a teacher trajectory. Together, these fields preserve the connection between a reviewed question, the resulting state and graph updates, the revised decision, and its evidence path.

ATLAS uses its structured consultation agents as a multi-agent teacher. The teacher processes the 39 development cases under three bounded review settings, $K\in\{1,2,3\}$, producing 117 structured trajectories. Each trajectory records internal safety questions, patient-state updates, PMCG transitions, risk assessments, component-level revisions, stopping decisions, and evidence paths. Policy distillation removes case-specific fields, merges recurring guideline-consistent transitions, resolves conflicting actions by risk priority, and stores the resulting symbolic policy in a versioned YAML rule table. The policy governs question selection, risk priority, medication reconciliation, targeted revision, stopping, and evidence verification. It contains no evaluation-case identifiers, test labels, or learned parameters. A deterministic compilation step converts the YAML policy into the frozen runtime representation consumed by ATLAS inference. The compiler preserves rule conditions, priorities, actions, and evidence requirements. All reported ATLAS experiments use this compiled frozen policy.

\begin{table*}[!t]
\caption{Teacher-trajectory schema used for symbolic policy distillation.}
\centering
\begin{small}
\begin{tabular}{@{}p{0.22\textwidth}p{0.68\textwidth}@{}}
\toprule
\textbf{Trajectory field} & \textbf{Description} \\
\midrule
\texttt{trajectory\_id} & Anonymous development trajectory identifier. \\
\texttt{review\_budget} & Bound on teacher-review rounds used to generate the development trajectory. \\
\texttt{initial\_state} & Structured state before review. \\
\texttt{questions} & Teacher-internal safety checks and decision-changing information needs. \\
\texttt{state\_updates} & Normalized evidence added during review. \\
\texttt{pmcg\_transitions} & Activated, resolved, or revised graph relations. \\
\texttt{risk\_assessments} & Conflict, geriatric-risk, caution, and monitoring judgments. \\
\texttt{decision\_revisions} & Component-level changes linked to graph updates. \\
\texttt{stopping\_decision} & Reason for terminating review. \\
\texttt{evidence\_paths} & Patient-fact and guideline support for the final claims. \\
\texttt{final\_decision} & Structured recommendation, avoidance, caution, alternative, unsafe, and trace fields. \\
\bottomrule
\end{tabular}
\end{small}

\label{tab:supp_c3}
\end{table*}

\subsection{C.5 Policy Compilation and Freeze Verification}

Table~\ref{tab:supp_c4} reports the release checks used to compare the symbolic source policy, compiled runtime representation, and frozen experimental artifact. These checks are intended to detect implementation drift rather than to add new clinical rules after development.

The distillation pipeline writes the symbolic policy to a versioned YAML file. A deterministic compiler converts this source policy into the frozen JSON representation read by the inference engine. The compilation process does not introduce new clinical rules or case-specific conditions. The release records the source-policy hash, compiled-policy hash, compiler version, rule count, and compilation command. We verify that the compiled artifact preserves every rule condition, priority, action, and evidence requirement. We also compare the compiled policy with the frozen policy used for the reported experiments through structural comparison and case-level behavioral testing. The verification requires identical structured outputs for the frozen and compiled policies on the development cases and all released ATLAS evaluation inputs. A mismatch causes the verification script to fail.

\begin{table*}[!t]
\centering
\caption{Policy-distillation and freeze-verification checks.}
\label{tab:supp_c4}
\begin{small}
\begin{tabular}{@{}p{0.31\textwidth}p{0.59\textwidth}@{}}
\toprule
\textbf{Verification item} & \textbf{Required result} \\
\midrule
Development trajectories & 117 \\
Development cases & 39 \\
Review settings & $K\in\{1,2,3\}$ \\
Case-specific conditions in policy & None \\
Test-label access & None \\
YAML-to-runtime compilation & Deterministic \\
Rule conditions and priorities & Exact match \\
Case-level outputs & Identical \\
Policy used for reported experiments & Compiled frozen policy \\
\bottomrule
\end{tabular}
\end{small}
\end{table*}

\subsection{C.6 Inference Procedure}

Algorithm 1 presents the complete inference order from state initialization to verified output. The procedure makes explicit that graph updates precede localized revision and that the trace verifier operates on the reconciled medication sets.

\begin{figure*}[!t]
\centering

\begin{minipage}{0.94\textwidth}
\begingroup
\fboxsep=7pt
\fboxrule=0.5pt

\fcolorbox{gray!55}{gray!10}{%
\begin{minipage}{
  \dimexpr\linewidth-2\fboxsep-2\fboxrule\relax
}
\footnotesize

\noindent
\textbf{Algorithm 1: ATLAS inference procedure.}\par
\smallskip

\noindent
\textbf{Input:} Visible patient state $s_0$, candidate set
$\mathcal{C}$, guideline graph $G$, compiled frozen policy $\Pi$,
and query budget $K$.\par
\smallskip

\begin{tabularx}{\linewidth}{
  @{}r@{\hspace{0.7em}}X@{}
}
1: &
Normalize the input and initialize a provisional PMCG
$G^{\mathrm{p}}_0$. \\

2: &
\textbf{for} $t=0,\ldots,K$ \textbf{do} \\

3: &
\hspace*{1em}Audit contraindications, geriatric risks, cautions,
and alternatives. \\

4: &
\hspace*{1em}Identify unresolved relations that can change the
medication plan. \\

5: &
\hspace*{1em}\textbf{if} no informative unresolved relation remains
\textbf{then break}. \\

6: &
\hspace*{1em}Select the highest-priority question $q_t$. \\

7: &
\hspace*{1em}Receive answer $a_t$ and update
$s_{t+1}=s_t\oplus\Phi(q_t,a_t)$. \\

8: &
\hspace*{1em}Rebuild the affected PMCG relations and revise
dependent decisions. \\

9: &
\textbf{end for} \\

10: &
Reconcile avoidance, caution, recommendation, and alternative
components. \\

11: &
Verify evidence paths and apply the Safety Gate. \\

12: &
\textbf{return} the structured medication-safety decision and trace. \\
\end{tabularx}

\end{minipage}%
}

\endgroup
\end{minipage}
\end{figure*}

\section{D. GeriMedBench Environment and Evaluation}
\label{app:section-d}
This section describes the interactive environment independently of any particular agent. Table~\ref{tab:supp_d1} defines the information boundary, Table~\ref{tab:supp_d2} defines the evaluation dimensions, and Table~\ref{tab:supp_d3} connects pipeline failures to their observable consequences.
\renewcommand{\thetable}{D\arabic{table}}
\setcounter{table}{0}

\subsection{D.1 Environment State and Interaction Loop}

Table~\ref{tab:supp_d1} identifies which environment objects are visible to the agent and which remain evaluator-only. Algorithm 2 complements this boundary by showing the action loop under the fixed question budget.

GeriMedBench separates the initial public state from hidden safety-critical facts. The environment returns only the answer to the agent's current question. It does not expose unrelated hidden facts. The interaction ends when the agent returns a final decision or uses all $K=3$ questions.

\begin{table*}[!t]
\caption{GeriMedBench environment objects and access boundary.}
\centering
\begin{small}
\begin{tabular}{@{}p{0.19\textwidth}p{0.49\textwidth}p{0.21\textwidth}@{}}
\toprule
\textbf{Environment object} & \textbf{Contents} & \textbf{Access rule} \\
\midrule
Public case & Presenting problem, therapeutic goal, known conditions, current medications, candidates, and visible context. & Visible to the agent at the start. \\
Hidden state & Safety-critical facts that may alter avoidance, caution, monitoring, or alternative decisions. & Accessible only to the environment. \\
Response map & Question-to-answer mapping grounded in the hidden state. & Returns one answer for the current question. \\
Interaction trace & Questions, answers, state updates, and decision revisions. & Recorded for evaluation. \\
Reference decision & Guideline-grounded final medication components and trace requirements. & Frozen evaluator only. \\
\bottomrule
\end{tabular}
\end{small}

\label{tab:supp_d1}
\end{table*}

\begin{figure*}[!t]
\centering

\begin{minipage}{0.94\textwidth}
\begingroup
\fboxsep=7pt
\fboxrule=0.5pt

\fcolorbox{gray!55}{gray!10}{%
\begin{minipage}{
  \dimexpr\linewidth-2\fboxsep-2\fboxrule\relax
}
\footnotesize

\noindent
\textbf{Algorithm 2: Interactive evaluation loop used by GeriMedBench.}\par
\smallskip

\noindent
\textbf{Input:} Public state $s_0$, hidden state $h$, and query budget
$K=3$.\par
\smallskip

\begin{tabularx}{\linewidth}{
  @{}r@{\hspace{0.7em}}X@{}
}
1: &
$t\leftarrow 0$. \\

2: &
\textbf{while} $t<K$ and the agent has not returned a final decision
\textbf{do} \\

3: &
\hspace*{1em}Agent selects either
\textsc{Ask}(question) or \textsc{Final}(decision). \\

4: &
\hspace*{1em}\textbf{if} the action is \textsc{Ask} \textbf{then} \\

5: &
\hspace*{2em}Environment returns only
$\operatorname{answer}(h,\text{question})$. \\

6: &
\hspace*{2em}Agent updates the public state with the revealed
evidence. \\

7: &
\hspace*{1em}\textbf{end if} \\

8: &
\hspace*{1em}$t\leftarrow t+1$. \\

9: &
\textbf{end while} \\

10: &
Evaluate the final decision, revision behavior, and evidence trace. \\
\end{tabularx}

\end{minipage}%
}

\endgroup
\end{minipage}
\end{figure*}

\subsection{D.2 Interactive Metrics}

Table~\ref{tab:supp_d2} defines the interactive metrics at the level of observable behavior. Reading the dimensions together distinguishes acquiring useful evidence from merely producing a strong final answer without an evidence-consistent revision path.

\begin{table*}[!t]
\caption{GeriMedBench evaluation dimensions.}
\centering
\begin{small}
\begin{tabular}{@{}p{0.17\textwidth}p{0.41\textwidth}p{0.31\textwidth}@{}}
\toprule
\textbf{Metric} & \textbf{What it measures} & \textbf{Interpretation} \\
\midrule
Information Gain & Whether the selected questions reveal safety-relevant hidden information. & Higher values indicate that the interaction acquires more decision-relevant evidence. \\
Query Efficiency & Safety-relevant information obtained relative to the limited question budget. & Higher values indicate that fewer or more focused questions obtain useful evidence. \\
Revision Accuracy & Whether newly revealed evidence produces the required component-level decision change. & A high score requires both correct acquisition and correct downstream revision. \\
Final Strict & Joint correctness of the final structured decision after interaction. & A single missing or unsupported component causes case-level failure. \\
Trace Consistency & Consistency between the interaction trace, acquired facts, and final claims. & Penalizes unsupported claims and revisions disconnected from new evidence. \\
Agent OSRS & Overall safety reasoning across final components, unsafe outcomes, and trace quality. & Summarizes safety reasoning on a 0--100 scale. \\
\bottomrule
\end{tabular}
\end{small}

\label{tab:supp_d2}
\end{table*}

\subsection{D.3 Interactive Failure Taxonomy}

Table~\ref{tab:supp_d3} assigns each failure stage an observable error and a downstream consequence. This stage-wise view supports error analysis without treating all unsuccessful consultations as equivalent.

The interactive pipeline can fail before or after a relevant fact is acquired. The error taxonomy separates these stages so that a low Final Strict score is not reduced to final-answer accuracy alone.

\begin{table*}[!t]
\caption{Error taxonomy for the interactive pipeline.}
\centering
\begin{small}
\begin{tabular}{@{}p{0.18\textwidth}p{0.39\textwidth}p{0.32\textwidth}@{}}
\toprule
\textbf{Failure stage} & \textbf{Observable error} & \textbf{Downstream consequence} \\
\midrule
Question selection & The agent asks about a low-impact fact while a high-risk dependency remains unresolved. & Hidden safety evidence remains unavailable. \\
State grounding & The response is misparsed, negated, or assigned to the wrong patient-state field. & The PMCG reflects an incorrect patient state. \\
PMCG update & A newly revealed fact does not activate or resolve the expected graph relation. & The relevant audit is not triggered. \\
Risk classification & The system identifies the fact but assigns the wrong avoidance or caution level. & Unsafe recommendation or incomplete monitoring. \\
Revision & The agent acquires the right fact but fails to update the affected decision component. & Low Revision Accuracy and Final Strict. \\
Evidence alignment & The final claim does not follow from the acquired evidence path. & Trace inconsistency. \\
Schema handling & The output is incomplete or cannot be parsed into the required structure. & Case-level failure under strict evaluation. \\
\bottomrule
\end{tabular}
\end{small}

\label{tab:supp_d3}
\end{table*}

\section{E. Baseline Configuration and Output Handling}
\label{app:section-e}
This section records the comparison boundary used by the offline evaluator. Table~\ref{tab:supp_e1} groups the compared systems, and Table~\ref{tab:supp_e2} lists the shared inputs, decoding controls, parsing rules, and reference-access restrictions.
\renewcommand{\thetable}{E\arabic{table}}
\setcounter{table}{0}

\subsection{E.1 Compared Systems}

Table~\ref{tab:supp_e1} organizes the baselines by system type so that foundation-model, retrieval-augmented, multi-agent, and proposed-system results can be interpreted consistently.

\begin{table*}[!t]
\caption{Method categories used in the reported comparisons.}
\centering
\begin{small}
\begin{tabular}{@{}p{0.20\textwidth}p{0.70\textwidth}@{}}
\toprule
\textbf{Category} & \textbf{Methods} \\
\midrule
Foundation models & Mistral-Small-3.2-24B; Qwen3-30B-A3B; DeepSeek-R1-Distill-Qwen-32B; MedGemma-27B-Text; GPT-5; Claude Opus 4.6; Gemini 3.1 Pro Preview \\
Knowledge-augmented models & Llama-3.3-70B-Instruct with BM25 retrieval; RotatE \\
Multi-agent system & MDAgents \\
Proposed system & ATLAS \\
\bottomrule
\end{tabular}
\end{small}

\label{tab:supp_e1}
\end{table*}

\subsection{E.2 Shared Evaluation Settings}

Table~\ref{tab:supp_e2} records the controls shared within each comparison protocol. The common parser and failure handling are especially important because malformed output must not be confused with a clinically correct structured decision.

\begin{table*}[!t]
\caption{Shared baseline settings and fairness controls.}
\centering
\begin{small}
\begin{tabular}{@{}p{0.25\textwidth}p{0.65\textwidth}@{}}
\toprule
\textbf{Setting} & \textbf{Value or rule} \\
\midrule
Visible input & Identical case fields and candidate set for all compared methods. \\
Output constraint & Shared structured JSON instruction and common target schema. \\
Temperature & 0.0 for LLM-based baselines. \\
Maximum generation length & 4096 tokens for LLM-based baselines. \\
Local serving context & vLLM \texttt{max\_model\_len} = 8192 for local model servers. \\
Retrieval & BM25 \texttt{top\_k} = 8 for the retrieval-augmented baseline. \\
Parser & Shared robust JSON parser for LLM, RAG, and multi-agent outputs. \\
Evaluation references & Unavailable to all inference runners; readable only by the frozen offline evaluator. \\
Prompt tuning & No prompt tuning on evaluation cases. \\
Invalid output & Handled by the same runner-level retry and failure policy within each shared comparison protocol. \\
\bottomrule
\end{tabular}
\end{small}

\label{tab:supp_e2}
\end{table*}

\subsection{E.3 Structured Output Handling}

The parser applies the schema introduced in Section I without consulting evaluation references. Consequently, schema validation and clinical scoring remain separate stages of the evaluation pipeline.

The parser extracts the first valid JSON object, normalizes aliases, and validates the required fields. A response that remains invalid after the shared runner policy receives the common failure treatment. The evaluator does not recover missing medication components from free text.

Fairness principle. The comparison fixes visible information, candidate medications, output schema, parser, and failure rules. It evaluates complete systems and does not isolate the independent contribution of guideline access, graph personalization, or multi-agent control.

\section{F. Additional Quantitative Results}
\label{app:section-f}
This section provides numerical detail beyond the compact summaries in the main paper. Tables~\ref{tab:supp_f1}--\ref{tab:supp_f4} report expanded static results, interactive acquisition metrics, single-disease generalization, and component-level ablation effects.
\renewcommand{\thetable}{F\arabic{table}}
\setcounter{table}{0}

\subsection{F.1 Expanded Western Results for Proprietary Baselines and ATLAS}

Table~\ref{tab:supp_f1} expands the Western comparison with component-level scores, trace pass rate, unsafe rate, and OSRS. The additional columns show whether a method's aggregate result reflects balanced medication reasoning or strength in only a subset of components.

Table~\ref{tab:supp_f1} adds $M_{\mathrm{avoid}}$ F1 and Trace Pass Rate to the proprietary-model comparison. These metrics clarify why high avoidance recall alone does not guarantee a complete medication decision. ATLAS does not lead every component. Its advantage lies in joint completion: high recommendation quality, complete avoidance recall, stronger caution and alternative coverage, and no unsafe recommendation under the automated evaluator occur in the same structured output.

\begin{table*}[!t]
\caption{Expanded results on the Western Multimorbidity Evaluation Set ($N=201$). Values are percentages except OSRS.}
\centering
\begin{scriptsize}
\begin{tabular*}{\textwidth}{@{\extracolsep{\fill}}lccccccccc@{}}
\toprule
\textbf{Method}
& \shortstack{\textbf{Strict}\\\textbf{Success}}
& \shortstack{$\boldsymbol{M_{\mathrm{rec}}}$\\\textbf{F1}}
& \shortstack{$\boldsymbol{M_{\mathrm{avoid}}}$\\\textbf{Recall}}
& \shortstack{$\boldsymbol{M_{\mathrm{avoid}}}$\\\textbf{F1}}
& \shortstack{$\boldsymbol{M_{\mathrm{caution}}}$\\\textbf{F1}}
& \shortstack{$\boldsymbol{M_{\mathrm{alt}}}$\\\textbf{F1}}
& \textbf{Unsafe}
& \textbf{Trace}
& \textbf{OSRS} \\
\midrule
GPT-5 & 32.34 & 69.15 & 99.00 & 90.05 & 41.29 & 49.12 & 0.50 & 100.00 & 71.37 \\
Claude Opus 4.6 & 34.83 & 88.56 & 98.01 & 97.35 & 46.77 & 85.32 & 0.00 & 100.00 & 82.30 \\
Gemini 3.1 Pro Preview & 38.31 & 63.68 & 99.00 & 97.84 & 70.65 & 80.60 & 0.00 & 100.00 & 82.58 \\
\rowcolor{gray!15}
ATLAS & 92.04 & 92.50 & 100.00 & 91.57 & 75.47 & 92.84 & 0.00 & 92.04 & 97.21 \\
\bottomrule
\end{tabular*}
\end{scriptsize}

\label{tab:supp_f1}
\end{table*}

\subsection{F.2 GeriMedBench Acquisition Metrics}

Table~\ref{tab:supp_f2} places acquisition and revision measures beside final decision quality. This joint presentation is necessary because interactive success depends on both obtaining decision-changing evidence and using it correctly.

\begin{table*}[!t]
\caption{ATLAS acquisition and final-decision metrics on GeriMedBench ($N=76$, $K=3$). Values are percentages except Agent OSRS.}
\centering
\begin{small}
\begin{tabular*}{\textwidth}{@{\extracolsep{\fill}}lccccccc@{}}
\toprule
\textbf{Method} & \shortstack{\textbf{Info}\\\textbf{Gain}} & \shortstack{\textbf{Query}\\\textbf{Efficiency}} & \shortstack{\textbf{Revision}\\\textbf{Accuracy}} & \shortstack{\textbf{Final}\\\textbf{Strict}} & \shortstack{\textbf{Unsafe}\\\textbf{Rate}} & \shortstack{\textbf{Trace}\\\textbf{Consistency}} & \shortstack{\textbf{Agent}\\\textbf{OSRS}} \\
\midrule
\rowcolor{gray!15}
ATLAS & 73.90 & 35.75 & 44.17 & 23.68 & 0.00 & 85.53 & 64.09 \\
\bottomrule
\end{tabular*}
\end{small}

\label{tab:supp_f2}
\end{table*}

\subsection{F.3 Tabular Single-Disease Comparison}

Table~\ref{tab:supp_f3} provides the numerical values underlying the main-paper visualization. Reporting the values directly makes the small differences in Strict Success and OSRS independently inspectable.

\begin{table*}[!t]
\caption{Numerical values corresponding to Figure 6 in the main paper. Strict Success is a percentage; OSRS uses a 0--100 scale.}
\centering
\begin{small}
\begin{tabular*}{\textwidth}{@{\extracolsep{\fill}}lcc@{}}
\toprule
\textbf{Method} & \textbf{Strict Success Rate} & \textbf{OSRS} \\
\midrule
GPT-5 & 76.47 & 94.77 \\
Claude Opus 4.6 & 81.21 & 95.18 \\
Mistral-Small-3.2-24B & 53.27 & 91.21 \\
Qwen3-30B-A3B & 79.41 & 95.82 \\
DeepSeek-R1-Distill-Qwen-32B & 77.94 & 96.03 \\
MedGemma-27B-Text & 78.59 & 96.36 \\
Gemini 3.1 Pro Preview & 92.60 & 97.84 \\
Llama-3.3-70B-Instruct & 56.05 & 90.73 \\
RotatE & 0.00 & 30.69 \\
MDAgents & 64.38 & 83.20 \\
\rowcolor{gray!15}
ATLAS & 94.12 & 97.55 \\
\bottomrule
\end{tabular*}
\end{small}

\label{tab:supp_f3}
\end{table*}

Under this evaluation, ATLAS leads Strict Success by 1.52 points over Gemini 3.1 Pro Preview, while Gemini leads OSRS by 0.29 points. The two metrics capture different properties, so the comparison supports stronger complete-decision performance rather than uniform superiority on every aggregate score.

\subsection{F.4 Metric-Specific Ablation Effects}

Table~\ref{tab:supp_f4} relates each ablation to its primary observed effect. The comparison is component-specific because removing a module need not affect recommendation, avoidance, caution, alternative, trace, and unsafe outcomes equally.

\begin{table*}[!t]
\caption{Key ablation effects reported in the main-text analysis.}
\centering
\begin{footnotesize}
\begin{tabularx}{\textwidth}{@{}p{0.14\textwidth}cccccX@{}}
\toprule
\textbf{Variant}
& \textbf{Strict}
& \shortstack{$\boldsymbol{M_{\mathrm{avoid}}}$\\\textbf{Recall}}
& \shortstack{$\boldsymbol{M_{\mathrm{caution}}}$\\\textbf{F1}}
& \shortstack{\textbf{Unsafe}\\\textbf{Rate}}
& \textbf{OSRS}
& \textbf{Primary effect} \\
\midrule
\rowcolor{gray!15}
Full ATLAS & 92.04 & 100.00 & 75.47 & 0.00 & 97.21 & Reference system \\
w/o PMCG & 19.90 & 48.26 & 40.30 & 51.74 & 48.78 & Removes patient-specific graph personalization and sharply degrades both safety and completeness. \\
w/o Geriatric Risk Auditor & 38.81 & 93.53 & 40.30 & 6.47 & 80.47 & Weakens caution and monitoring decisions. \\
w/o Drug Conflict Auditor & 69.15 & 69.15 & 59.70 & 30.85 & 74.41 & Misses medication--condition conflicts and avoidance decisions. \\
w/o Safety Gate & 26.87 & 73.13 & 40.30 & 26.87 & 65.90 & Allows unresolved or inconsistent decisions to reach the final output. \\
\bottomrule
\end{tabularx}
\end{footnotesize}

\label{tab:supp_f4}
\end{table*}

\section{G. Qualitative Case Study and Error Taxonomy}
\label{app:section-g}
This section follows individual evidence changes through the PMCG and final medication decision. Tables~\ref{tab:supp_g1} and~\ref{tab:supp_g2} provide concrete revision traces, while Table~\ref{tab:supp_g3} supplies a common template for assigning observed failures to pipeline stages.
\renewcommand{\thetable}{G\arabic{table}}
\setcounter{table}{0}

\subsection{G.1 Illustrative Graph-Guided Revision}

Table~\ref{tab:supp_g1} traces the PMCG changes triggered by newly revealed evidence. Example G1 presents the corresponding structured medication sets and evidence paths after revision.

The following case study expands the knee-pain example in Figure~1 of the main paper. It is an illustrative trace used to explain the mechanism, not an additional benchmark result.

\begin{table*}[!t]
\caption{Illustrative PMCG update and decision revision for the knee-pain example.}
\centering
\begin{small}
\begin{tabular}{@{}p{0.13\textwidth}p{0.27\textwidth}p{0.26\textwidth}p{0.23\textwidth}@{}}
\toprule
\textbf{Step} & \textbf{Observed evidence} & \textbf{PMCG state} & \textbf{Decision consequence} \\
\midrule
Initial request & Older adult asks what to take for chronic knee pain. Hypertension is known. Kidney disease is unreported. & NSAID-related renal-risk dependency remains unresolved. & A one-shot recommendation would be premature. \\
Targeted question & ATLAS asks about kidney disease or renal function because the answer may change the safety of NSAIDs. & The unresolved renal-risk relation receives highest priority. & The system delays the final plan. \\
New evidence & The response reveals chronic kidney disease. & CKD activates a medication--condition conflict for NSAIDs. & NSAIDs move to $M_{\mathrm{avoid}}$. \\
Alternative search & Guideline evidence supports acetaminophen within an appropriate use context. & Alternative relation links the excluded option to a safer candidate. & Acetaminophen enters $M_{\mathrm{rec}}$, while a distinct safer strategy or guideline-supported substitute enters $M_{\mathrm{alt}}$ when appropriate. \\
Caution and monitoring & Blood pressure, pain response, dose, and follow-up remain clinically relevant. & Monitoring dependencies attach to the final plan. & $M_{\mathrm{caution}}$ records monitoring and use conditions. \\
Verification & Trace Verifier links the avoidance and alternative claims to CKD and guideline evidence. & Safety Gate finds no unresolved conflict. & ATLAS returns the structured plan. \\
\bottomrule
\end{tabular}
\end{small}

\label{tab:supp_g1}
\end{table*}

\begin{table*}[!t]
\centering
\begin{minipage}{0.94\textwidth}
\textbf{Example G1: Presentation of structured decision-making and evidence pathways.}
\begin{small}
\begin{description}
  \item[$M_{\mathrm{rec}}$:] Acetaminophen within an appropriate use context.
  \item[$M_{\mathrm{avoid}}$:] NSAIDs, including ibuprofen, because CKD increases kidney-injury risk.
  \item[$M_{\mathrm{caution}}$:] Monitor dose, blood pressure, pain response, and follow-up needs.
  \item[$M_{\mathrm{alt}}$:] Non-pharmacologic pain management or another guideline-supported option when appropriate.
  \item[$U$:] No unsafe recommendation.
  \item[Evidence path:] CKD $\rightarrow$ renal-risk conflict $\rightarrow$ NSAID avoidance $\rightarrow$ safer alternative search.
\end{description}
\end{small}
\end{minipage}
\end{table*}

\subsection{G.2 Evaluated GeriMedBench Trace}

Table~\ref{tab:supp_g2} shows a released consultation trace with the observed evidence, ATLAS action, and decision effect aligned by stage. The alignment makes it possible to verify that each revision follows the information actually revealed at that point.

Table~\ref{tab:supp_g2} follows one released GeriMedBench case from its visible state through three questions to the final medication report. The case exposes both a success and an inefficiency: the first two questions retrieve the decisive renal and gastrointestinal facts, whereas the third question adds no new risk information.

\begin{table*}[!t]
\caption{Released GeriMedBench trace for a 79-year-old patient seeking analgesia. The case is guideline-derived and contains no identifiable patient information.}
\centering
\begin{small}
\begin{tabular}{@{}p{0.15\textwidth}p{0.30\textwidth}p{0.26\textwidth}p{0.18\textwidth}@{}}
\toprule
\textbf{Stage} & \textbf{Evidence available} & \textbf{ATLAS action} & \textbf{Decision effect} \\
\midrule
Initial state & Age 79; analgesia need; aspirin 325 mg or nonselective NSAIDs under review; acetaminophen and nondrug support are visible candidates. & Marks renal and gastrointestinal safety dependencies as unresolved. & Delays the final plan rather than treating the initial profile as complete. \\
Query 1: renal function & The environment reveals renal-risk context relevant to NSAID use and dose accumulation. & Updates the patient state and activates the renal-risk conflict. & Strengthens the avoidance case for aspirin and nonselective NSAIDs. \\
Query 2: gastrointestinal history & The environment reveals gastric or acid-peptic disease with increased ulcer and bleeding risk. & Adds the gastrointestinal conflict to the PMCG and re-audits the candidate set. & Confirms that the high-risk option should move to $M_{\mathrm{avoid}}$. \\
Query 3: heart-failure status & No additional heart-failure decompensation is documented. & Closes a remaining safety branch without activating a new conflict. & No medication component changes. \\
Final report & The two critical facts are available and linked to the medication decision. & Recommends acetaminophen, avoids aspirin or nonselective NSAIDs, returns a safe alternative, and passes the Safety Gate. & The case satisfies the complete structured reference under the evaluator. \\
\bottomrule
\end{tabular}
\end{small}

\label{tab:supp_g2}
\end{table*}

The trace shows why GeriMedBench separates information acquisition from final-decision quality. ATLAS obtains both critical facts and returns the complete reference-aligned decision, yet the third question does not improve the plan. The case contributes positively to Final Strict while leaving room for better Query Efficiency.

\subsection{G.3 Why One-Shot Accuracy Is Insufficient}

The evaluated traces illustrate why the final answer alone cannot reveal whether a system asked an informative question, updated the state correctly, or revised the affected medication component for the right reason.

The case separates answer plausibility from decision validity. Ibuprofen may appear reasonable when the initial message is treated as a complete patient profile. The renal condition changes the decision. ATLAS succeeds only if it identifies the missing dependency, asks a focused question, updates the graph, revises the affected medication components, and preserves a trace from evidence to action.

\subsection{G.4 Error-Analysis Template}

Table~\ref{tab:supp_g3} standardizes the evidence recorded for each error category. Using the same reviewer question and remediation target helps distinguish state-grounding failures from PMCG, risk, revision, evidence, and schema failures.

\begin{table*}[!t]
\caption{Trace-oriented template for qualitative error analysis.}
\centering
\begin{small}
\begin{tabular}{@{}p{0.18\textwidth}p{0.37\textwidth}p{0.34\textwidth}@{}}
\toprule
\textbf{Error category} & \textbf{Reviewer question} & \textbf{Recommended evidence to inspect} \\
\midrule
Missed information & Did the system leave a decision-changing fact unresolved? & Question ranking and unresolved PMCG relations. \\
Patient-state error & Did the system parse the answer into the correct state field? & Grounded state before and after the response. \\
PMCG error & Did the new fact activate or resolve the expected relation? & Graph transition record. \\
Risk-classification error & Was the fact assigned to avoidance, caution, or monitoring correctly? & Auditor output and guideline relation. \\
Alternative error & Did the substitute remain safe under the updated patient state? & Alternative relation and revised safety audit. \\
Revision error & Did every affected component change after the new evidence? & Before/after structured decision diff. \\
Evidence-alignment error & Does each final claim follow from a visible or acquired fact? & Evidence path and trace-verification output. \\
Schema error & Did the final output satisfy the required structured format? & Parser report and missing-field record. \\
\bottomrule
\end{tabular}
\end{small}

\label{tab:supp_g3}
\end{table*}

\section{H. Blinded Clinician Evaluation}
\label{app:section-h}
This section separates the clinician-review protocol from its aggregate outcomes. Table~\ref{tab:supp_h1} defines the review dimensions, and Tables~\ref{tab:supp_h2} and~\ref{tab:supp_h3} report overall and benchmark-stratified ratings.
\renewcommand{\thetable}{H\arabic{table}}
\setcounter{table}{0}

\subsection{H.1 Review Design}

Table~\ref{tab:supp_h1} defines the blinded review dimensions before any scores are summarized. This definition separates clinical correctness and safety from completeness, actionability, evidence consistency, and overall preference.

Three clinicians independently reviewed 40 benchmark-stratified cases, with 20 cases from the Western evaluation and 20 from GeriMedBench. Each reviewer compared one ATLAS output with one Gemini 3.1 Pro Preview output. We randomized case order and A/B assignment for each reviewer, removed model identifiers and automated scores, and presented both outputs through the same structured template.

\begin{table*}[!t]
\caption{Clinician-review dimensions.}
\centering
\begin{small}
\begin{tabular}{@{}p{0.23\textwidth}p{0.67\textwidth}@{}}
\toprule
\textbf{Review item} & \textbf{Definition} \\
\midrule
Clinical correctness & Whether the medication decision is clinically appropriate for the presented patient state. \\
Medication safety & Whether the plan avoids harmful options and states necessary safety constraints. \\
Decision completeness & Whether the output covers recommendation, avoidance, caution, and alternative needs. \\
Actionability & Whether the plan gives a usable next step, monitoring action, or follow-up instruction. \\
Evidence consistency & Whether the rationale matches the patient evidence and guideline-grounded decision. \\
Potentially unsafe & Case-level judgment that an output contains a potentially unsafe recommendation. \\
Preferred output & Reviewer preference between the two blinded outputs, with ties allowed. \\
\bottomrule
\end{tabular}
\end{small}

\label{tab:supp_h1}
\end{table*}

\subsection{H.2 Overall and Benchmark-Stratified Results}

Table~\ref{tab:supp_h2} reports the aggregate clinician ratings and unsafe-case judgments. Table~\ref{tab:supp_h3} separates the Western and GeriMedBench subsets to show whether the overall pattern is concentrated in one benchmark.

\begin{table*}[!t]
\caption{Overall blinded clinician evaluation. Scores use a five-point scale and average all reviewer--case ratings. Unsafe counts use case-level majority judgments. The remaining 7 cases were rated as ties.}
\centering
\begin{small}
\begin{tabular*}{\textwidth}{@{\extracolsep{\fill}}lccccccc@{}}
\toprule
\textbf{Method} & \textbf{Correct} & \textbf{Safety} & \textbf{Complete} & \textbf{Action} & \textbf{Evidence} & \textbf{Unsafe} & \textbf{Preferred} \\
\midrule
Gemini 3.1 Pro Preview & 3.77 & 4.12 & 3.67 & 3.85 & 3.80 & 2/40 & 5/40 \\
\rowcolor{gray!15}
ATLAS & 4.11 & 4.35 & 4.01 & 4.01 & 4.08 & 1/40 & 28/40 \\
\bottomrule
\end{tabular*}
\end{small}

\label{tab:supp_h2}
\end{table*}

\begin{table*}[!t]
\caption{Benchmark-stratified mean clinician ratings.}
\centering
\begin{small}
\begin{tabular*}{\textwidth}{@{\extracolsep{\fill}}llccccc@{}}
\toprule
\textbf{Benchmark} & \textbf{Method} & \textbf{Correct} & \textbf{Safety} & \textbf{Complete} & \textbf{Action} & \textbf{Evidence} \\
\midrule
Western & Gemini 3.1 Pro Preview & 3.97 & 4.30 & 3.90 & 4.02 & 4.02 \\
Western & ATLAS & 4.45 & 4.60 & 4.40 & 4.22 & 4.37 \\
GeriMedBench & Gemini 3.1 Pro Preview & 3.57 & 3.95 & 3.43 & 3.68 & 3.58 \\
GeriMedBench & ATLAS & 3.77 & 4.10 & 3.62 & 3.80 & 3.78 \\
\bottomrule
\end{tabular*}
\end{small}

\label{tab:supp_h3}
\end{table*}

ATLAS receives higher mean ratings in all five criteria in both benchmark strata. The separation is larger on the Western cases, where the automated evaluation also shows stronger complete-decision performance. The reviewed sample remains small, and the results do not establish real-world clinical effectiveness.

\subsection{H.3 Agreement and Interpretation}

The agreement analysis should be read together with Tables~\ref{tab:supp_h2} and~\ref{tab:supp_h3}. The clinician study is a blinded assessment of the sampled outputs and is not evidence of prospective clinical effectiveness.

Ordinal Krippendorff's alpha ranges from 0.33 to 0.71 across the five criteria. The variation indicates that agreement depends on the criterion and should be interpreted with the small sample and ordinal rating scale in mind.

\section{I. Shared Prompt Templates and Output Schema}
\label{app:section-i}
This section reproduces the shared task-facing instructions needed to
interpret parser behavior and invalid outputs. Prompt I1 covers static
decisions, Prompt I2 defines interactive actions, and Schema I1 gives
the common structured medication-safety output.

\subsection{I.1 Shared Static-Baseline Prompt}

Prompt I1 shows the common static task instruction, including the
visible patient state, candidate medications, and required structured
decision fields.

\begin{figure*}[!t]
\centering

\begin{minipage}{0.94\textwidth}
\begingroup
\fboxsep=7pt
\fboxrule=0.5pt

\fcolorbox{gray!55}{gray!10}{%
\begin{minipage}{
  \dimexpr\linewidth-2\fboxsep-2\fboxrule\relax
}
\footnotesize
\raggedright

\noindent
\textbf{Prompt I1: Shared static-baseline instruction template.}
\par\smallskip

{\ttfamily
System role: You are a medication-safety decision-support model
for older adults.
\par\smallskip

Input:
\par
- Patient state: \{visible\_patient\_state\}
\par
- Candidate medications: \{candidate\_set\}
\par
- Task: produce a complete structured medication-safety decision.
\par\smallskip

Requirements:
\par
1. Recommend only supported options.
\par
2. Identify medications that should be avoided.
\par
3. State cautions, monitoring, dose, or duration conditions.
\par
4. Provide safer alternatives when an option is excluded.
\par
5. Do not assume that missing information is absent.
\par
6. Return valid JSON only and follow the required schema.
\par
7. Do not mention or infer evaluation labels.
\par
}

\end{minipage}%
}

\endgroup
\end{minipage}
\end{figure*}

\subsection{I.2 GeriMedBench Action Format}

Prompt I2 restricts each interactive response to either one question
or one final structured decision, which keeps environment transitions
machine-checkable.

\begin{figure*}[!t]
\centering

\begin{minipage}{0.94\textwidth}
\begingroup
\fboxsep=7pt
\fboxrule=0.5pt

\fcolorbox{gray!55}{gray!10}{%
\begin{minipage}{
  \dimexpr\linewidth-2\fboxsep-2\fboxrule\relax
}
\footnotesize
\raggedright

\noindent
\textbf{Prompt I2: Interactive action contract for GeriMedBench.}
\par\smallskip

{\ttfamily
At each turn, return exactly one action:
\par\smallskip

ASK:
\par
\{
\par
\hspace*{1em}"action": "ask",
\par
\hspace*{1em}"question":
"\textless one safety-critical question\textgreater"
\par
\}
\par\smallskip

or FINAL:
\par
\{
\par
\hspace*{1em}"action": "final",
\par
\hspace*{1em}"decision":
\{ ... structured medication decision ... \}
\par
\}
\par\smallskip

You may ask at most K = 3 questions. Base each question only on the
visible state and previously revealed answers.
\par
}

\end{minipage}%
}

\endgroup
\end{minipage}
\end{figure*}

\subsection{I.3 Structured Decision Schema}

Schema I1 defines the medication sets, monitoring conditions, unsafe
flag, and evidence paths consumed by the shared parser and evaluator.

\begin{figure*}[!t]
\centering

\begin{minipage}{0.94\textwidth}
\begingroup
\fboxsep=7pt
\fboxrule=0.5pt

\fcolorbox{gray!55}{gray!10}{%
\begin{minipage}{
  \dimexpr\linewidth-2\fboxsep-2\fboxrule\relax
}
\footnotesize
\raggedright

\noindent
\textbf{Schema I1: The structured medication-safety output.}
\par\smallskip

{\scriptsize\ttfamily
\{
\par
\hspace*{1em}"M\_rec": [
\par
\hspace*{2em}\{"medication": "...", "use\_context": "...",
"evidence\_path": ["..."]\}
\par
\hspace*{1em}],
\par
\hspace*{1em}"M\_avoid": [
\par
\hspace*{2em}\{"medication": "...", "reason": "...",
"evidence\_path": ["..."]\}
\par
\hspace*{1em}],
\par
\hspace*{1em}"M\_caution": [
\par
\hspace*{2em}\{"medication": "...",
"monitoring\_or\_condition": "...",
"evidence\_path": ["..."]\}
\par
\hspace*{1em}],
\par
\hspace*{1em}"M\_alt": [
\par
\hspace*{2em}\{"medication\_or\_strategy": "...",
"rationale": "...", "evidence\_path": ["..."]\}
\par
\hspace*{1em}],
\par
\hspace*{1em}"M\_level": "...",
\par
\hspace*{1em}"U": false,
\par
\hspace*{1em}"sources": ["..."]
\par
\}
\par
}

\end{minipage}%
}

\endgroup
\end{minipage}
\end{figure*}

\subsection{I.4 Parsing and Invalid-Output Handling}

The invalid-output policy applies the same structural requirements
described in Schema I1. A response is scored only after parsing and
normalization produce the expected fields without introducing
information absent from the model output.

The shared parser extracts the first valid JSON object, normalizes
equivalent field names, and checks the required components. It does
not fill missing clinical content from free text. Outputs that remain
invalid after the shared runner policy receive the common failure
treatment defined by the evaluation protocol.

\end{document}